\pdfoutput=1

\documentclass[11pt]{article}

\usepackage{acl}

\usepackage{times}
\usepackage{latexsym}

\usepackage[T1]{fontenc}

\usepackage[utf8]{inputenc}

\usepackage{microtype}

\usepackage{inconsolata}
\usepackage{algorithm}  
\usepackage{algorithmic}
\usepackage{wrapfig}
\usepackage{nicefrac}
\usepackage{booktabs}       
\usepackage{amsfonts}
\usepackage{xcolor}

\usepackage{amsthm,amsmath,bm,xspace}
\usepackage{bbm}
\usepackage{color}
\usepackage{enumitem}
\usepackage{breqn}
\usepackage{pifont}
\usepackage{soul}
\usepackage{multirow}
\usepackage[capitalize,noabbrev]{cleveref}
\usepackage{amssymb}
\usepackage{subcaption}
\usepackage{caption}

\usepackage{amsmath}
\DeclareMathOperator*{\argmax}{arg\,max}

\DeclareSymbolFont{extraup}{U}{zavm}{m}{n}
\DeclareMathSymbol{\varheart}{\mathalpha}{extraup}{86}
\DeclareMathSymbol{\vardiamond}{\mathalpha}{extraup}{87}

\def\eg{{\em e.g.,}\xspace}
\def\ie{{\em i.e.,}\xspace}

\def\Plus{\texttt{+}}
\def\Minus{\texttt{-}}

\def\std#1{\ \tiny{$\pm{#1}$}}

\def\xicl{X-ICL\xspace}

\def\caseone{ICL\xspace}
\def\casetwo{X-ICL (Human)\xspace}
\def\casethree{fs-X-ICL (ChatGPT)\xspace}
\def\casefour{zs-X-ICL (ChatGPT)\xspace}
\def\casefive{fs-X-ICL (Llama2)\xspace}
\def\casesix{fs-X-ICL(Vicuna)\xspace}

\def\caseeight{zs-X-ICL (ChatGPT$_{\text{s}}$)\xspace}
\def\caseswap{fs-X-ICL (ChatGPT$_{\text{swap}}$)\xspace}
\def\caserand{X-ICL (Human$_{\text{rand}}$)\xspace}

\def\cosim{\textsc{COSINE}\xspace}
\def\bms{\textsc{BM25}\xspace}
\def\bsr{\textsc{SET-BSR}\xspace}

\def\figref#1{Figure~\ref{#1}}

\def\tabref#1{Table~\ref{#1}}

\def\secref#1{section~\ref{#1}}

\def\eqref#1{(\ref{#1})}

\def\1{\bm{1}}

\def\vr{{\bm{r}}}
\def\vs{{\bm{s}}}

\def\vx{{\bm{x}}}

\def\vy{{\bm{y}}}

\usepackage{marginnote}
\usepackage{xcolor}
\usepackage{soul}

\newcommand{\expspace}{\mathcal{E}}
\newcommand{\instspace}{\mathcal{X}}
\newcommand{\labelspace}{\mathcal{Y}}

\definecolor{myred}{RGB}{215,48,39}
\definecolor{mygreen}{RGB}{26,152,80}

\title{Using Natural Language Explanations to Improve \\ Robustness of In-context Learning}

\author{Xuanli He$^\clubsuit$ \qquad Yuxiang Wu$^\varheart$ \qquad Oana-Maria Camburu$^\clubsuit$ \\ \textbf{Pasquale Minervini}$^\spadesuit$ \qquad \textbf{Pontus Stenetorp}$^\clubsuit$ \\
$^\clubsuit$University College London \qquad $^\varheart$Weco AI \qquad $^\spadesuit$University of Edinburgh\\
\texttt{\small{z.xuanli.he@gmail.com \quad yuxiang@weco.ai }}\quad \texttt{\small{p.minervini@ed.ac.uk}} \\ \texttt{\small{\{o.camburu, p.stenetorp\}@ucl.ac.uk}}  \\
}

\begin{document}
\maketitle
\begin{abstract}
Recent studies demonstrated that large language models~(LLMs) can excel in many tasks via in-context learning~(ICL).
However, recent works show that ICL-prompted models tend to produce inaccurate results when presented with adversarial inputs.
In this work, we investigate whether augmenting ICL with natural language explanations (NLEs)
improves the robustness of LLMs on adversarial datasets covering natural language inference and paraphrasing identification.
We prompt LLMs with a small set of human-generated NLEs to produce further NLEs, yielding more accurate results than both a zero-shot-ICL setting and using only human-generated NLEs.
Our results on five popular LLMs~(GPT3.5-turbo, Llama2, Vicuna, Zephyr, and Mistral) show that our approach 
yields over 6\% improvement over baseline approaches for eight adversarial datasets: HANS, ISCS, NaN, ST, PICD, PISP, ANLI, and PAWS.
Furthermore, previous studies have demonstrated that prompt selection strategies significantly enhance ICL on in-distribution test sets.
However, our findings reveal that these strategies do not match the efficacy of our approach for robustness evaluations, resulting in an accuracy drop of 8\%  compared to the proposed approach.\footnote{Code and datasets are accessible at: \url{https://github.com/xlhex/acl2024_xicl}}
%
%
\end{abstract}

\section{Introduction}
The landscape of AI has recently undergone a significant transformation with the advent of large language models (LLMs).
These models can produce accurate predictions on unseen data after observing a small number of demonstrations.
Remarkably, they can achieve this based on examples provided directly in their inputs, without explicit retraining or fine-tuning -- this learning paradigm is referred to as \emph{in-context learning}~\cite[ICL,][]{NEURIPS2020_1457c0d6,rae2021scaling}.
However, ICL struggles to execute complex tasks, such as arithmetic, commonsense, and symbolic reasoning~\cite{rae2021scaling}.
To improve the effectiveness of ICL in solving tasks requiring complex reasoning, \citet{wei2022chain} drew inspiration from
natural language explanations~(NLEs) to introduce a method denoted as the Chain-of-Thought~(CoT) prompting.
%
%
CoT prompting involves prompting a model with a sequence of intermediate steps or reasoning processes to guide it towards generating more accurate answers.\footnote{CoTs and NLEs are similar concepts, as they both describe the reasoning process behind a decision in natural language; as NLEs were introduced before CoTs~\citep{camburu2018snli,Hendricks_2018_ECCV},
we use the former term.}
%
%
%
In this work, we denote ICL equipped with NLEs as \textit{\xicl}.
Despite its simplicity, \xicl has advanced the performance of ICL across a broad range of complex reasoning tasks~\cite{wei2022chain, wang2023selfconsistency}.
Similarly to supervised learning, ICL tends to be vulnerable to adversarial examples~\cite{wang2023adversarial}.
Previous research shows that improving the robustness of fine-tuned models against such adversarial datasets is possible by fine-tuning with task-relevant NLEs~\cite{chen2022can,ludan-etal-2023-explanation}.
Inspired by this, we hypothesize that incorporating NLEs into ICL could also improve the robustness of LLMs against adversarial examples.
To this end, we evaluate the robustness of \xicl on eight adversarial datasets: HANS, ISCS, NaN, ST, PICD, PISP, ANLI, and PAWS.
%
%

\begin{figure}[t]
    \centering
    \includegraphics[width=0.85\linewidth]{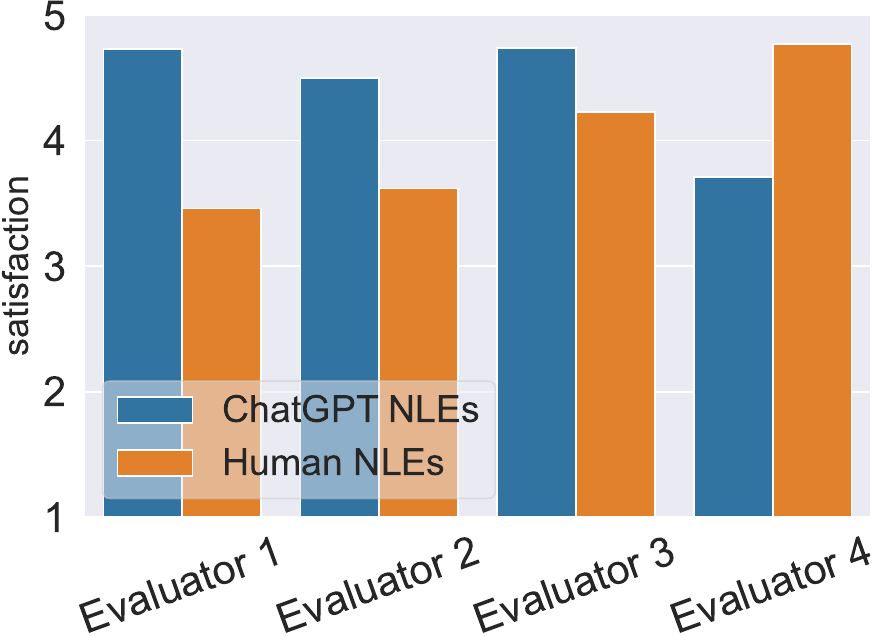}
    \caption{Human evaluation on 100 NLEs generated by GPT3.5-turbo (labeled as \emph{ChatGPT NLEs}) and 100 NLEs generated by human annotators (labeled as \emph{Human NLEs}). The satisfaction scores span from 1~(extremely dissatisfied) to 5~(extremely satisfied).}
    \label{fig:eval_nles}
    \vspace{-0.4cm}
\end{figure}

%
Moreover, the effectiveness of \xicl so far relies on the availability of human-written NLEs~\cite{wei2022chain}, 
which usually require domain-specific knowledge, making them hard to collect.
However, the advent of LLMs uncovered a range of possibilities where LLMs can assist human annotators~\citep{DBLP:journals/corr/abs-2302-04023,DBLP:journals/corr/abs-2301-07597}.
Motivated by this development, we investigate using three LLMs, namely GPT3.5-turbo, Llama2, and Vicuna, to generate NLEs for ICL.
%
We then use human annotators to assess the quality of 200 human-written and LLM-generated NLEs.
As shown in \cref{fig:eval_nles}, most annotators~(3 out of 4) prefer NLEs produced by ChatGPT (GPT3.5-turbo) over those crafted by humans.\footnote{More details are available in Appendix~\ref{app:qual_analysis_demo}.} This observation further motivates us to 
evaluate models prompted with LLM-generated NLEs.

We then evaluate the improvement in the robustness of \xicl in three settings -- in two of the settings, an LLM is prompted with LLM-generated NLEs (generated in zero-shot-ICL and few-shot-ICL settings, and in the last setting, the LLM is prompted with human-generated NLEs.
%
%
In the evaluation, we consider five popular LLMs (\ie Mistral~\cite{jiang2023mistral}, Zephyr~\cite{tunstall2023zephyr},  Vicuna~\cite{vicuna2023}, Llama2~\cite{touvron2023llama} and GPT3.5-turbo) on eight adversarial datasets.
Our experimental results suggest that \xicl produces more accurate results than ICL and, moreover, that NLEs generated by ChatGPT in a few-shot-ICL setting (by prompting ChatGPT with human-generated NLEs) significantly improve over the ICL baseline~(+6\%) for the majority of the considered datasets and LLMs.
%
%
%
Thus, our findings suggest that an integrated approach, combining human inputs with LLMs, can provide a more effective solution than utilizing either human annotators or LLMs in isolation. 
Finally, we show that while prompt selection strategies (\ie retrieving relevant training examples) 
can significantly improve the accuracy of ICL on in-distribution test sets \cite{gupta2023coverage,levy-etal-2023-diverse,10.5555/3618408.3620070}, they are less effective on adversarial datasets when compared to X-ICL methods, with our approach~(few-shot-ICL) outperforming them by more than 8\% in accuracy.

\section{Related Work}
\paragraph{Learning with Explanations.} 
There has been a surge of work on explaining predictions of neural NLP systems, from highlighting decision words \cite{10.1145/2939672.2939778, alvarez-melis-jaakkola-2017-causal, serrano-smith-2019-attention} to generating NLEs~\cite{camburu2018snli, narang2020wt5, teachmetoexplain}. Our work concentrates on the latter category, namely, the self-generation of NLEs for justifying model predictions. \citet{rajani-etal-2019-explain} propose a two-stage training process to improve the prediction performance for commonsense reasoning tasks. In their work, the first stage revolves around generating NLEs, which are then used to inform the label prediction training process in the second stage. Alternatively, one can leverage a multi-task framework to generate NLEs and labels simultaneously~\cite{hase-etal-2020-leakage}. \citet{li2022explanations} propose advancing the reasoning abilities of smaller LMs by leveraging NLEs generated by GPT-3~\cite{NEURIPS2020_1457c0d6}.
NLEs have also vastly been employed beyond NLP, such as in computer vision \citep{Hendricks_2018_ECCV, zellers_recognition_2019, rexc}, in the medical domain \citep{mimicnle}, and for self-driving cars \citep{cars}, with some works showing improved task performance when training with NLEs \cite{kayser2021vil}. However, these studies primarily concentrate on supervised fine-tuning approaches, which is different from the focus of this work, \ie \caseone.

\paragraph{Prompting with NLEs.}
%
Despite its remarkable performance on several downstream tasks~\cite{NEURIPS2020_1457c0d6},
ICL 
can still produce inaccurate results in tasks requiring reasoning abilities, such as arithmetic, logical, and commonsense reasoning tasks~\cite{rae2021scaling,srivastava2022beyond}.
To improve the reasoning abilities of LLMs, \citet{wei2022chain} introduced CoT prompting.
This technique prompts an LM to generate a sequence of concise sentences that imitate the reasoning process an individual might undergo to solve a task before providing the ultimate answer, essentially to provide an NLE/CoT before generating the final answer.
Furthermore, \citet{wang2023selfconsistency} propose to improve CoT prompting by combining multiple diverse reasoning paths generated by LLMs, enhancing the accuracy of a greedy CoT prompting approach. 
However, these aforementioned methods need human-written NLEs as CoT in the prompts.
Instead, our LLM-based zero-shot-ICL regime harnesses the power of an LLM to synthesize NLEs without human-written NLEs.
%


\begin{figure*}[t!]
    \centering
    \includegraphics[width=\textwidth]{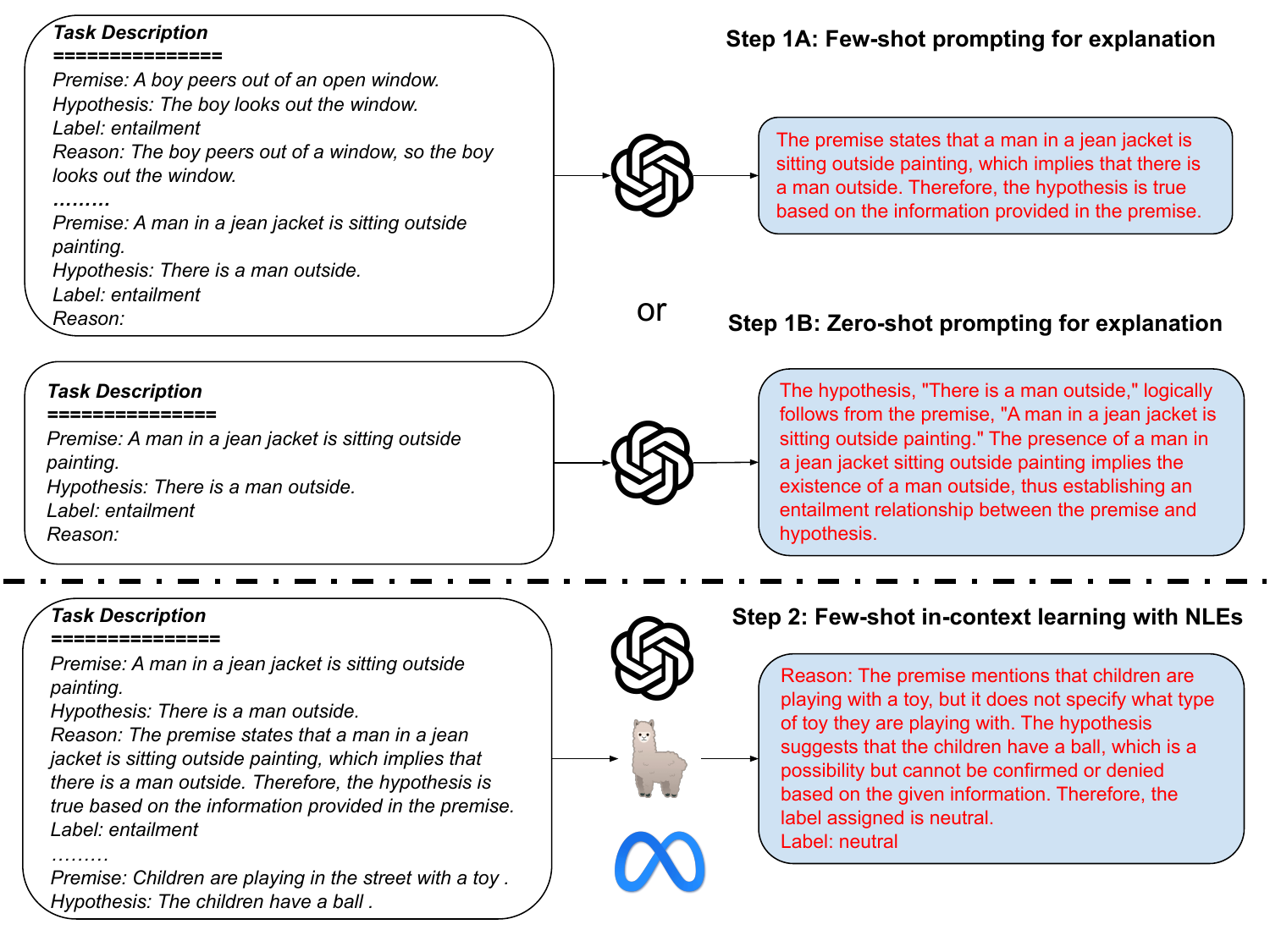}
    \caption{Illustrction of using LLM-generated NLEs for ICL: (1) prompt an LLM in a few-shot or zero-shot manner to generate NLEs for new instances; (2) prompt LLMs using \caseone with the NLEs generated in step 1. }
    \label{fig:workflow}
\end{figure*}

\paragraph{Learning Robust Models.}
%
Several works show that NLP models are prone to performance degradation when presented with adversarial examples, a consequence of inherent artifacts or biases within the annotation of the training dataset~\cite{naik-etal-2018-stress, mccoy-etal-2019-right, nie-etal-2020-adversarial, liu-etal-2020-empirical}. Various strategies have been proposed to mitigate biases within NLP models, \eg initially training a weak model to recognize superficial features, subsequently enforcing a target model to learn more robust and generalizable characteristics~\cite{he-etal-2019-unlearn,clark-etal-2019-dont,karimi-mahabadi-etal-2020-end, yaghoobzadeh-etal-2021-increasing, korakakis-vlachos-2023-improving}.
Additionally, data augmentation presents another viable option~\cite{minervini-riedel-2018-adversarially,wu-etal-2021-polyjuice, wu-etal-2022-generating}. Moreover, %
studies have shown that supervised fine-tuning of models using rationales or human-written NLEs can significantly enhance the models' resilience against adversarial datasets~\cite{chen2022can,stacey2022supervising, kavumba-etal-2023-prompting, ludan-etal-2023-explanation}. Unlike them, our research examines the robustness of X-ICL across eight adversarial datasets, highlighting a novel finding: NLEs generated by LLMs surpass those produced by human annotators in enhancing model robustness. In addition, unlike human-written NLEs, those produced by LLMs exhibit greater scalability and adaptability across diverse tasks.

\section{Methodology}
\label{sec:method}
This section first outlines the workflow of \xicl. Then, the focus shifts to detailing how an LLM can generate an NLE for a labeled instance.

\subsection{\caseone with NLEs (\xicl)}
\label{sec:cot}
LLMs can provide significantly more accurate predictions across various reasoning tasks when supplied with human-written NLEs~\cite{wei2022chain,wei2022emergent}.
In \xicl, given an instance, the task is to generate the most likely prediction and NLE for that instance.
More formally, in \xicl, given an unlabeled instance $\vx^\prime \in \instspace$ and a set of training examples $(\vx_i, \vr_i, \vy_i)$, where $\vx_i \in \instspace$ is an instance, $\vy_i \in \labelspace$ is its label, and $\vr_i \in \expspace$ is the corresponding explanation, the task is to identify the most likely label and explanation for $\vx^\prime$:
\begin{equation*}
\argmax_{(\vr',\vy')\in \expspace \times \labelspace} P_{\theta}\left( (\vr^\prime, \vy^\prime) \mid (\vx_i,\vr_i,\vy_i)_{i=1}^k,(\vx') \right),
\end{equation*}
\noindent where $\theta$ denotes the model parameters, and $\instspace$, $\labelspace$, and $\expspace$ are the sets of all possible instances, labels, and explanations, respectively.
%

%
%
The objective is to generate the most likely combination of label $\vy^\prime$ and explanation $\vr^\prime$ from an LLM, after prompting it with the demonstration examples, including labeled instances and NLEs $(\vx_i,\vr_i,\vy_i)_{i=1}^k$, as well as the unlabeled instance $\vx^\prime$.

\subsection{Generating NLEs with LLMs}
\label{sec:explanation_gen}
In existing \xicl works, human-written NLEs $\vr$ were used for the instances within the demonstration set. Instead, in this work, we opt for the NLEs synthesized via LLMs. This preference is driven by noting that NLEs produced by LLMs tend to receive higher approval ratings from human evaluators, as indicated in~\figref{fig:eval_nles}. We argue that this preference will boost the performance of \xicl. The methods utilized for the generation of NLEs are outlined below.

\paragraph{Few-shot prompting for NLEs} Our methodology, also shown in \cref{fig:workflow}, initiates by leveraging a set of labeled instances, each accompanied by a human-crafted NLE, to prompt LLMs. The primary aim is to encourage the LLMs to generate a correct NLE (\ie the ground-truth arguments) for the correctly predicted answer for a test instance.
The most likely NLE is then generated as follows:
\begin{equation} \label{eq:fewshot}
\argmax_{\vr' \in \expspace } P_{\theta}(\vr' \mid \vs,(\vx_j,\vy_j,\vr_j)_{j=1}^m, (\vx',\vy')),
\end{equation}
\noindent where $\vs$ denotes a meta-prompt representing the task. More details on the meta-prompt and demonstration sets are available in Appendix~\ref{app:meta_prompt}.

\paragraph{Zero-shot prompting for NLEs} We further extend our approach to situations where human-written NLEs are absent, which is generally more prevalent across most datasets. In this context, LLMs are prompted to generate an NLE for a labeled instance devoid of any pre-existing examples with NLEs. The objective bears a resemblance to Equation (\ref{eq:fewshot}), albeit without the inclusion of the demonstration set $(\vx_j,\vy_j,\vr_j)_{j=1}^m$.

Notably, the NLEs generated by the aforementioned approaches can be seamlessly integrated into the existing \xicl framework as delineated in \cref{sec:cot}.
We primarily focus on using GPT-3.5 (more specifically, GPT3.5-turbo-0613 -- we will refer to this model as ChatGPT) to synthesize NLEs. Given that LLMs, such as ChatGPT, may have been trained on datasets incorporating NLEs, it challenges the assumption of genuine zero- or few-shot learning scenarios. To clarify terminology and avoid confusion, we redefine `zero-shot learning' as the absence of demonstration sets, and `few-shot ICL' as learning that utilizes a demonstration set. Thus, we denote the aforementioned two approaches as \casefour and \casethree, respectively. In addition, we explore the application of two other widely used open-source LLMs for generating NLEs. Detailed results of these experiments are provided in Appendix~\ref{app:suppl}.

\section{Experiments}
We conduct a series of experiments to assess the performance of our proposed \xicl framework.

\subsection{Experimental Setup}
\paragraph{Tasks and datasets} We consider the Natural Language Inference (NLI) and paraphrasing identification tasks as our testbed. To ascertain the robustness of LLMs when employing the proposed approach, we evaluate it across eight adversarial datasets.
For the NLI task, we include HANS, ISCS, ST, PICD, PISP, NaN, and ANLI. The first five datasets (HANS, ISCS, ST, PICD, PISP) are from \citet{liu-etal-2020-empirical}, while NaN and ANLI are sourced from \citet{truong-etal-2022-another} and \citet{nie-etal-2020-adversarial}, respectively. Regarding the paraphrasing identification task, we use the PAWS-QQP (or PAWS) dataset~\cite{zhang-etal-2019-paws}.

Additionally, the SNLI dataset~\cite{bowman-etal-2015-large} and QQP~\cite{wang-etal-2018-glue}, which are non-adversarial, are employed for a comparative purpose. The details of these datasets are provided in Appendix~\ref{app:data}.

\begin{table*}[!th]
    \centering
    \scalebox{0.77}{
    \begin{tabular}{clcccccccc|cc| c}
    \toprule
    \multirow{2}{*}{\textbf{Models}}&  \multirow{2}{*}{\textbf{Methods}}& \multicolumn{8}{c|}{\textbf{Natural Language Inference}} & \multicolumn{2}{c|}{\textbf{Paraphrasing}} & \multirow{2}{*}{\textbf{Avg.}}\\
    \cmidrule{3-12}
    & & \textbf{SNLI} & \textbf{HANS} & \textbf{ISCS}& \textbf{NaN}	& \textbf{ST} & \textbf{PICD} & \textbf{PISP} & \textbf{ANLI} & \textbf{QQP} & \textbf{PAWS}  \\
    \midrule
   \multirow{7}{*}{{\rotatebox[origin=c]{90}{\textbf{Mistral 7B}}}}& \caseone &59.8\ \ \ & 54.0\ \ \  & 51.9\ \ \  & 55.0\ \ \  & 44.4\ \ \  & 58.2\ \ \  & 23.0\ \ \  &  39.8\ \ \ & 69.9\ \ \ & 68.3\ \ \ & 50.3\vspace{-0.15cm}\\
    & & \std{3.4} & \std{2.2} & \std{1.4} & \std{1.3} & \std{1.7} & \std{2.6} & \std{2.6} & \std{4.6} & \std{1.7}& \std{2.7}\\
   &\casetwo & 60.0\ \ \  & 56.0\ \ \ 	& 54.7$^\triangledown$ & 58.6$^\triangledown$ & 51.7$^\blacktriangledown$  & 56.9\ \ \  & 35.8$^\blacktriangledown$  & 43.9$^\blacktriangledown$& 69.9\ \ \   & 66.4\ \ \ & 53.5\vspace{-0.15cm}\\
 & & \std{2.0} & \std{2.9} & \std{2.5} & \std{2.9} & \std{4.0} & \std{3.3} & \std{6.7} & \std{1.7} & \std{0.8} & \std{1.5}\\
   &\casefour & 56.7\ \ \  & 51.8\ \ \  & 47.7\ \ \  & 55.9\ \ \  & 44.9\ \ \  & 56.7\ \ \  & 25.1\ \ \  & 28.8\ \ \ & 67.3\ \ \  & 64.7\ \ \  & 46.4\vspace{-0.15cm}\\
 && \std{6.3} & \std{5.1} & \std{3.5} & \std{5.0} & \std{4.8} & \std{6.6} & \std{8.9} & \std{4.4} & \std{2.3} & \std{3.1}\\
   &\casethree & \textbf{61.8}\ \ \  & \textbf{58.2}$^\blacktriangledown$  & \textbf{57.2}$^\blacktriangledown$  & \textbf{62.4}$^\blacktriangledown$  & \textbf{55.2}$^\blacktriangledown$  & \textbf{59.2}\ \ \  & \textbf{47.6}$^\blacktriangledown$  & 	\textbf{46.9}$^\blacktriangledown$ & \textbf{70.3}\ \ \  & \textbf{72.5}$^\triangledown$ & \textbf{57.1}\vspace{-0.15cm}\\
    & &  \std{3.1} & \std{2.5} & \std{2.2} & \std{2.6} & \std{1.5} & \std{2.7} & \std{1.8} & \std{2.3} & \std{1.1} & \std{1.3}\\
   \midrule
\multirow{7}{*}{{\rotatebox[origin=c]{90}{\textbf{Zephyr 7B}}}} &\caseone &67.1\ \ \  & 71.0\ \ \  & 63.4\ \ \  & 65.7\ \ \  & 60.5\ \ \  & 64.8\ \ \  & 48.4\ \ \  & 47.1\ \ \ & 76.9\ \ \ & 57.7\ \ \	 & 59.8 \vspace{-0.15cm}\\
& &\std{3.4} & \std{1.8} & \std{1.2} & \std{1.8} & \std{1.0} & \std{1.5} & \std{1.4} & \std{1.6} & \std{0.4} & \std{1.1}\\
&\casetwo &72.4$^\blacktriangledown$& 64.3\ \ \  & 58.3\ \ \  & 62.0\ \ \  & 57.0\ \ \  & 60.6\ \ \  & 52.0\ \ \  & 49.4\ \ \ &75.8\ \ \ & 61.4$^\triangledown$	 & 59.3\vspace{-0.15cm}\\
& & \std{4.3} & \std{6.7} & \std{5.5} & \std{5.3} & \std{6.3} & \std{9.7} & \std{6.7} & \std{3.0} & \std{1.7} &	\std{2.3} \\
&\casefour& 67.2\ \ \  & 72.7\ \ \  & 60.4\ \ \  & 64.0\ \ \  & 61.4\ \ \  & 64.1\ \ \  & 50.8\ \ \  &40.9\ \ \ & 74.7\ \ \ & 59.1\ \ \ & 58.1\vspace{-0.15cm}\\
& & \std{3.9} & \std{2.6} & \std{5.3} & \std{5.2} & \std{5.7} & \std{5.4} & \std{5.2} & \std{3.8} & \std{1.8} & \std{2.4}\\
&\casethree & \textbf{74.2}$^\blacktriangledown$  & \textbf{77.4}$^\blacktriangledown$  & \textbf{67.0}\ \ \  & \textbf{67.7}\ \ \  & \textbf{69.3}$^\blacktriangledown$  & \textbf{70.0}$^\blacktriangledown$  & \textbf{65.6}$^\blacktriangledown$  & \textbf{52.1}$^\triangledown$ & \textbf{77.3}\ \ \ & \textbf{61.5}$^\triangledown$& \textbf{65.5}\vspace{-0.15cm}\\
&& \std{3.6} & \std{2.2} & \std{1.6} & \std{2.3} & \std{1.5} & \std{2.1} & \std{2.5} & \std{2.8} & \std{0.9} & \std{1.0}
\\
\midrule
\multirow{7}{*}{{\rotatebox[origin=c]{90}{\textbf{Vicuna 30B}}}}&\caseone& 65.2\ \ \ & 69.4\ \ \ &62.7\ \ \ & 61.4\ \ \ & 58.7\ \ \ & \textbf{67.1}\ \ \ & 50.9\ \ \ & 50.0\ \ \ & \textbf{81.8}\ \ \ & 69.7\ \ \ & 61.4\vspace{-0.15cm}\\
&&\std{2.7} &\std{1.2}	& \std{0.9} & \std{3.5} & \std{0.8} & \std{1.6} & \std{1.3} & \std{2.6} & \std{0.5} & \std{2.6}\\
&\casetwo & \textbf{67.8}\ \ \ & 62.9\ \ \ & 60.9\ \ \ & 64.2\ \ \ & 57.3\ \ \ & 63.7\ \ \ & 55.0\ \ \ &48.2\ \ \ & 77.4\ \ \ & 63.4\ \ \	& 59.8\vspace{-0.15cm}\\
&&\std{3.2}& \std{3.7}	 & \std{2.2} & \std{1.2} & \std{2.0} & \std{7.2} & \std{5.8} & \std{4.7} & \std{2.8} & \std{3.5}\\
&\casefour &64.2\ \ \ & 61.4\ \ \ & 64.9\ \ \ & 60.2\ \ \ & 61.7\ \ \ & 57.9\ \ \ & 51.8\ \ \ &49.7\ \ \ & 72.1\ \ \ & 61.8\ \ \	& 58.8\vspace{-0.15cm}\\
&&\std{5.9}& \std{7.7} & \std{2.3} & \std{4.0} & \std{3.1} & \std{8.7} & \std{8.7} & \std{3.6} & \std{3.2} & \std{4.9}\\
&\casethree & 65.0\ \ \  & \textbf{74.5}$^\triangledown$	& \textbf{65.5}$^\triangledown$	& \textbf{66.3}$^\triangledown$	& \textbf{64.8}$^\blacktriangledown$	& 61.6\ \ \ & \textbf{65.9}$^\blacktriangledown$	& \textbf{57.5}$^\blacktriangledown$ & 78.6\ \ \ & \textbf{70.0}\ \ \ & \textbf{65.4}\vspace{-0.15cm}\\
&&\std{3.1}& \std{4.4}	& \std{1.6} & \std{1.1} & \std{1.8} & \std{8.9} & \std{4.7} & \std{1.3} & \std{1.7} & \std{3.3}\\
\midrule
\multirow{7}{*}{{\rotatebox[origin=c]{90}{\textbf{Llama2 70B}}}}&\caseone&69.3\ \ \   & 65.7\ \ \  & \textbf{63.1}\ \ \  & 61.5\ \ \  & 58.8\ \ \  & 67.6\ \ \  & 48.5\ \ \  & 54.2\ \ \ & \textbf{80.8}\ \ \   & 44.5\ \ \ & 60.3\vspace{-0.15cm}\\
&&\std{1.2}& \std{3.4} & \std{1.6}	& \std{2.3} & \std{4.4} & \std{3.0} & \std{7.3}& \std{2.9} & \std{0.6}& \std{2.9}\\
&\casetwo& 73.0$^\blacktriangledown$  & 65.2\ \ \  & 59.6\ \ \  & 62.4\ \ \  & 55.7\ \ \  & 64.3\ \ \  & 50.4\ \ \  & 49.0\ \ \ &74.5\ \ \   & 42.6\ \ \   & 57.7\vspace{-0.15cm}\\
&& \std{3.1} &\std{4.6}	& \std{4.4}	& \std{3.3}	& \std{3.9} & \std{2.3} & \std{5.1} & \std{2.6} & \std{3.0} & \std{3.3} \\
&\casefour& 55.4\ \ \  & 64.0\ \ \  & 37.4\ \ \  & 58.1\ \ \  & 47.7\ \ \  & 53.5\ \ \  & 44.2\ \ \  & 35.8\ \ \ & 69.1\ \ \  & 37.8\ \ \  &	48.1\vspace{-0.15cm}\\
&&\std{5.5} &\std{6.3} & \std{6.0}	& \std{5.4} & \std{5.4} & \std{8.5} & \std{8.7} & \std{0.8} &\std{4.1} & \std{4.8} \\
&\casethree& \textbf{74.2}$^\blacktriangledown$  & \textbf{73.3}$^\blacktriangledown$  & 57.7\ \ \  & \textbf{65.9}$^\triangledown$ & \textbf{63.1}$^\triangledown$ & \textbf{70.6}$^\triangledown$ & \textbf{55.8}$^\blacktriangledown$  & \textbf{59.2}$^\blacktriangledown$ &  77.6\ \ \  & \textbf{46.5}$^\triangledown$ & \textbf{63.6}\vspace{-0.15cm}\\
&& \std{2.5} &\std{8.5} & \std{1.2} & \std{3.2} & \std{3.7} &\std{6.5} & \std{5.9} & \std{1.6} & \std{0.6} & \std{1.9} \\
\midrule
\multirow{7}{*}{{\rotatebox[origin=c]{90}{\textbf{GPT3.5-turbo}}}} &\caseone & 71.9\ \ \ &  72.4\ \ \ & 64.4\ \ \  	& 70.0\ \ \  & 62.1\ \ \ &64.0\ \ \   & 51.2\ \ \   &  56.1\ \ \ & \textbf{81.5}\ \ \ & 42.9\ \ \ & 62.4 \vspace{-0.15cm}\\
& & \std{1.4}& \std{0.6}	& \std{0.9} & \std{0.8} & \std{1.6} & \std{3.1} & \std{0.4} & \std{2.0}  & \std{0.3} & \std{2.8}\\
& \casetwo & \textbf{78.0}$^\blacktriangledown$& 71.0\ \ \  & 69.0$^\triangledown$ & 70.5\ \ \  & 65.7$^\triangledown$ & 72.7$^\blacktriangledown$& 59.3$^\triangledown$ & 59.8$^\triangledown$  & 76.0\ \ \ & 53.4$^\blacktriangledown$	& 66.2\vspace{-0.15cm}\\
&&\std{1.7}& \std{1.7}	&\std{1.2} & \std{2.2}& \std{1.0} & \std{1.3} & \std{1.9}	& \std{2.3} & \std{3.9} & \std{5.3}\\
&\casefour  &71.9\ \ \ & 71.6\ \ \  & 68.4$^\triangledown$	& 70.2\ \ \ 	& 67.6$^\triangledown$	& 67.7$^\triangledown$	& 61.7$^\blacktriangledown$	&\textbf{60.4}$^\blacktriangledown$  & 80.4\ \ \ & 51.2$^\blacktriangledown$	& 66.0\vspace{-0.15cm}\\
 & & \std{2.7}& \std{0.8}	& \std{0.3}	& \std{0.0}	&\std{1.3} & \std{4.1} & \std{1.9}	& \std{2.0} & \std{0.8} &	\std{3.1}\\
& \casethree &75.5$^\triangledown$& \textbf{76.0}$^\blacktriangledown$ & \textbf{74.9}$^\blacktriangledown$& \textbf{73.1}$^\blacktriangledown$ & \textbf{73.3}$^\blacktriangledown$ & \textbf{76.9}$^\blacktriangledown$ & \textbf{75.5}$^\blacktriangledown$	&59.6$^\triangledown$  & 79.0\ \ \ & \textbf{54.0}$^\blacktriangledown$	& \textbf{69.7}\vspace{-0.15cm}\\
&&\std{2.8}& \std{2.0} & \std{0.1} & \std{1.4} & \std{0.4} & \std{0.4} & \std{3.0} & \std{1.8} & \std{1.7} & \std{2.6}\\
    \bottomrule
    \end{tabular}
    }
    \caption{Accuracy of multiple LLMs using (1) standard ICL without NLEs, (2) \xicl with human-written NLEs: \casetwo, (3) \xicl with ChatGPT-generated NLEs in a zero-shot scenario: \casefour, (4)  \xicl with ChatGPT-generated NLEs in a few-shot scenario: \casethree. The best performance for each task within a model is shown in \textbf{bold}. Significance testing was assessed via an unequal variances \textit{t}-test in comparison with \caseone: $\blacktriangledown$ (resp. $\triangledown$) represents a \textit{p}-value lower than $10^{\texttt{-}3}$ (resp. $10^{\texttt{-}1}$). The results of ANLI are the average of ANLI R1, R2, and R3.}
    \label{tab:main}
    \vspace{-0.4cm}
\end{table*}

\paragraph{Language models and prompts} The evaluation of our approach is undertaken across five prominent LLMs: (1) Mistral, (2) Zephyr, (3) Vicuna, (4) Llama2, and (5) GPT3.5-turbo (version 0613). Specifically, the Mistral and Zephyr models have \texttt{7B} parameters each. For Vicuna and Llama2, we use the \texttt{30B} and \texttt{70B} versions, respectively.

We perform all \xicl experiments in an 8-shot setting, wherein each experiment is conducted four times independently, thereby drawing 32 unique instances from the training-associated datasets as follows. 
Specifically, for NLI datasets (except ANLI, which includes its own training set and NLEs), we adhere to the established methodology of using the e-SNLI dataset as the demonstration set, as suggested by~\citet{liu-etal-2020-empirical}.
The e-SNLI dataset is a modified version of SNLI, where each instance is annotated with NLEs written by humans.
In the case of the QQP and PAWS datasets, the QQP dataset is utilized as the demonstration set.
As no NLEs are available, we contribute the corresponding NLEs (refer to Appendix \ref{app:qqp}).
Regarding the generation of NLEs via few-shot learning described in \secref{sec:explanation_gen}, the methodology involves selecting a random instance from each label category within the training dataset to form the demonstration set. 
Consequently, the demonstration set comprises three instances for the e-SNLI dataset and two for the QQP dataset.

\paragraph{Baselines} In addition to the proposed method, our study investigates two baselines for comparative analysis. The first baseline uses standard ICL without NLEs. The second employs human-written NLEs within the \xicl process, referred to as \casetwo.

\subsection{Main Results}
This section examines \caseone and \xicl across the studied datasets using Mistral, Zephyr, Vicuna, Llama2, and GPT3.5-turbo. The results are summarized in~\tabref{tab:main}.

The results demonstrate a consistent outcome across both scenarios: with and without the application of \xicl. As the capabilities of the models increase, there is a noticeable improvement in average accuracy. This progression is evident when comparing the least potent model, exemplified by Mistral, to the most advanced one, represented by GPT3.5-turbo.

\tabref{tab:main} demonstrates that \casetwo yields better predictive accuracy than \caseone across all five LLMs assessed using the SNLI dataset, with enhancements of up to 6.1\%. This performance elevation is, however, limited to the Mistral and GPT-3.5-turbo models when subjected to all adversarial NLI test sets. The advantage of \casetwo relative to \caseone diminishes when applied to the QQP and PAWS datasets. 

For \casethree, both Mistral and Zephyr demonstrate a significant performance advantage in all evaluated tasks, outperforming \caseone and \casetwo by at least 5.7\% and 3.6\%, respectively. Despite the notable improvement on \caseone when employing GPT3.5-turbo in comparison to other LLMs, \casethree offers substantially additional gains, with an increase in absolute accuracy between 11\%-24\% on tasks such as ISCS, ST, PICD, PISP, and PAWS. This suggests that \xicl enhances LLM effectiveness on in-distribution test sets and increases their robustness against adversarial test sets.


Remarkably, despite the predominant preference of human evaluators for NLEs generated by GPT3.5 over those written by humans, \casefour consistently produces less accurate results than \casetwo across all models under study. The exception to this trend is GPT3.5-turbo, where a tie is observed. Furthermore, it appears counter-intuitive that \casefour is outperformed by \caseone for 4 out of the 5 LLMs analyzed, especially on Llama2. We conduct a systematic analysis in \secref{sec:suppl} to understand this apparent discrepancy between human preferences and LLM performance.

In light of the encompassment of diverse robustness scenarios by the seven adversarial NLI datasets, our primary focus henceforth will be the examination of these NLI datasets.

\subsection{Impacts of NLEs}
Our research has demonstrated that using NLEs generated by GPT3.5 can substantially enhance the performance of \xicl. To provide a more comprehensive understanding of the NLEs' influence, we conducted two investigations, presented below.

\paragraph{Data selection vs. \xicl.} The effectiveness of ICL in LLMs is closely linked to the quality of demonstrations provided, as these demonstrations are critical for the model's ability to understand and address the test instances~\cite{pmlr-v139-zhao21c,liu-etal-2022-makes,lu-etal-2022-fantastically}. Consequently, considerable research has focused on developing data selection techniques to optimize the curation of ICL demonstrations from relevant candidate data pools, aiming to enhance their alignment with the test instances~\cite{gupta2023coverage,levy-etal-2023-diverse,10.5555/3618408.3620070}. While these approaches have proven to be highly effective on in-distribution test sets, their performance on adversarial test sets remains uncertain, as these sets have the potential to misguide the selection algorithms.

\begin{table}[]
    \centering
    \scalebox{0.8}{
    \begin{tabular}{clccc}
    \toprule
    \textbf{Models}& \textbf{Methods}&\textbf{SNLI} & \textbf{AdvNLI} & \textbf{$\Delta$}\\
    \midrule
     \multirow{5}{*}{{\rotatebox[origin=c]{90}{\textbf{Zephyr}}}} & \caseone  & 67.1 & 57.2 & \ \ \textbf{9.9}\\
     & \casethree &74.2	& \textbf{63.7}	& 10.5 \\
     & \cosim & 77.0	& 55.6	& 21.4\\
     & \bms & 70.1	& 53.7	& 16.4 \\
      & \bsr & \textbf{79.9}	& 59.7	& 20.2\\
      \midrule
    \multirow{5}{*}{{\rotatebox[origin=c]{90}{\textbf{GPT3.5-turbo}}}} & \caseone & 71.9 & 61.4 & 10.5\\
     & \casethree & 75.5 & \textbf{69.8} & \ \ \textbf{5.6}\\
     & \cosim &75.0 & 58.1 & 16.9\\
     & \bms &71.4 & 56.0 & 15.4\\
      & \bsr & \textbf{77.4} & 59.5	& 17.9\\
    \bottomrule
    \end{tabular}}
    \caption{Performance of ICL, \casethree and three data selection approaches on SNLI and AdvNLI (\ie seven adversarial test sets). $\Delta$ indicates the difference between SNLI and adversarial NLI test sets. We report the average performance over all adversarial test sets.}
    \label{tab:select}
\end{table}

In this context, we compare the performance of \casethree to three prevalent data selection techniques: \cosim, \bms, and \bsr. \cosim incorporates sentence embeddings~\cite{reimers-gurevych-2019-sentence} to identify the most relevant demonstrations for each test instance, while \bms employs the BM25 algorithm~\cite{SPARCKJONES2000779} for retrieving candidate demonstrations. \bsr utilizes BERTScore~\cite{Zhang*2020BERTScore:}, integrated with set theory, to ensure comprehensive information coverage and diversity within the selected instances~\cite{gupta2023coverage}. Note that these data selection techniques are designed to sift through the entirety of the training data to choose demonstrations, a computationally demanding and computationally expensive process for generating NLEs for the full dataset. Therefore, our analysis is confined to applying ICL to these methods. To facilitate a generic comparison with the in-distribution set, we consider the average performance across all adversarial NLI test sets.

{According to \tabref{tab:select}, as expected, the data selection approaches markedly enhance ICL performance on the SNLI dataset for all studied LLMs, with notable improvements observed in \bsr, achieving gains of up to 17.8\% over standard ICL. However, this pronounced advantage diminishes considerably on adversarial test sets, particularly for \cosim and \bms models, which are outperformed by \caseone across all tested LLMs. This discrepancy results in a marked disparity between the in-distribution and adversarial test sets, contrary to what is observed in \casethree. These results imply that current data selection approaches may be prone to overfitting on in-distribution tests, potentially leading to significant challenges in processing OOD and adversarial datasets due to their limited generalizability.}

\begin{figure}
    \centering
\includegraphics[width=0.98\linewidth]{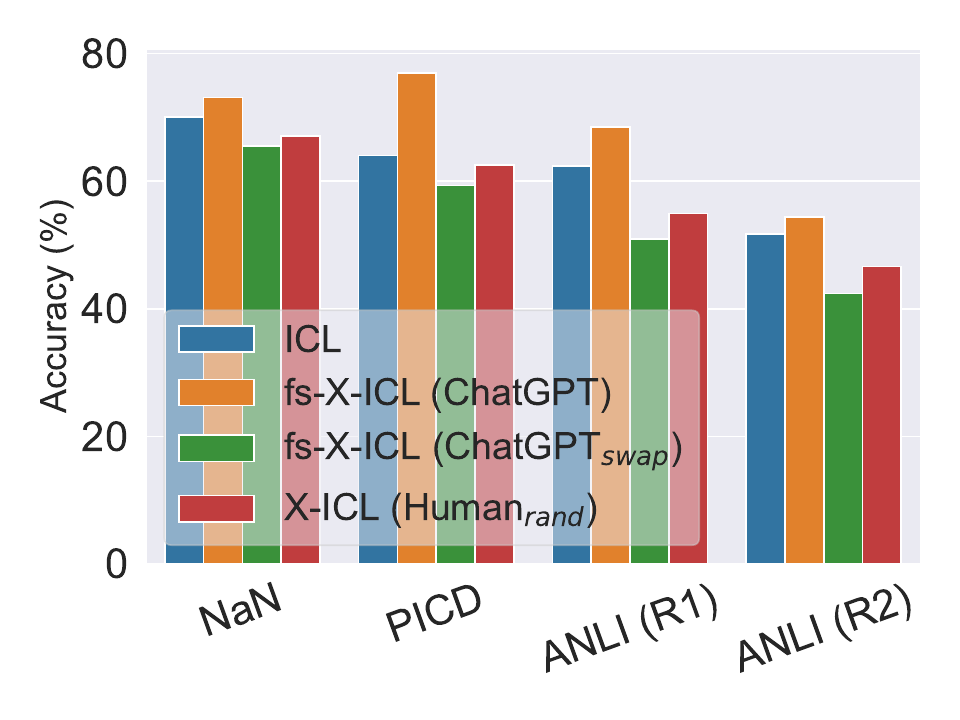}
    \caption{ICL performance of GPT3.5-turbo using (1) standard ICL without NLEs, (2) \xicl with GPT3.5-generated NLEs in a few-shot scenario: \casethree, (3) \xicl with GPT3.5-generated NLEs, where the NLEs of the prompt are swapped and do not match the instances: \caseswap, and (4) \xicl with random human NLEs: \caserand.}
    \label{fig:rand_exp}
\end{figure}

\paragraph{Do proper NLEs really help?} The prevailing assumption argues that the benefits of the \xicl primarily originate from the NLEs provided. To conclusively attribute these gains to the NLEs rather than any potential influence of additional sentences, we investigate two experimental setups. In the first setup, we randomly swap the NLEs within the prompt, leading to a mismatched NLE for each instance. This variant is henceforth referred to as \caseswap. Regarding the second variant, for each instance in the demonstration set, we randomly select an unrelated human NLE from the corresponding training set, referred to as \caserand.

As depicted in \figref{fig:rand_exp}, despite identical content being provided to GPT3.5-turbo, a misalignment between the NLE and the instance results in a marked reduction in the performance of \caseswap when compared to \casethree. This decline is discernible across various datasets, including NaN, PICD, and ANLI (R1/R2).\footnote{Similar patterns have been detected in other datasets} It is also shown that an irrelevant and arbitrary NLE triggers a performance reduction within the \xicl framework. Furthermore, the efficiency of both \caseswap and \caserand substantially lags behind that of \caseone. Therefore, it can be inferred that the efficacy of the \casethree hinges on providing an accurate and relevant NLE.

\begin{figure*}
    \centering
    \includegraphics[width=0.94\linewidth]{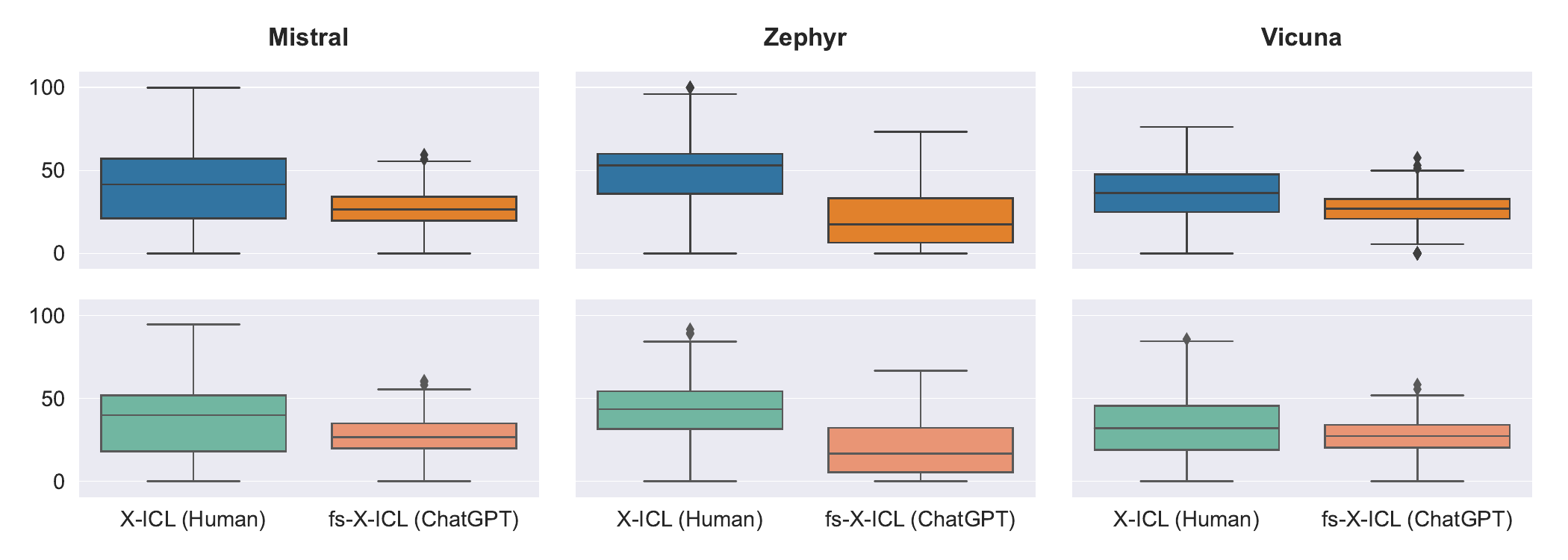}
    \caption{ROUGE-L between the NAN test set and the corresponding generated NLEs. \textbf{Top}: ROUGE-L between test premise and NLE. \textbf{Bottom}: ROUGE-L between test hypothesis and NLE.}
    \label{fig:rouge_nle}
    \vspace{-0.4cm}
\end{figure*}

\begin{table}[]
    \centering
   \scalebox{0.9}{
    \begin{tabular}{p{0.93\linewidth}}
    \toprule
    \textbf{Premise}: None of them supported her. \\
    \textbf{Hypothesis}: One of them supported her.
    \\
    \textbf{NLE [\casetwo]}: If none of them supported her, then one of them did not support her.\\
    \textbf{NLE [\casethree]}: The hypothesis contradicts the given premise, which states that none of them supported her.\\
    \midrule
    \textbf{Premise}: Not all people have had the opportunities you have had. \\
    \textbf{Hypothesis}: Some people have not had the opportunities you have had.
    \\
    \textbf{NLE [\casetwo]}: If not all people have had the opportunities you have had, then some people have not had the opportunities you have had.\\
    \textbf{NLE [\casethree]}: The hypothesis is a direct result of the premise, and the label assigned is entailment.\\
    \bottomrule
    \end{tabular}}
    \caption{Two test examples from the NAN dataset and the corresponding NLEs generated by \casetwo and \casethree using Zephyr.}
    \label{tab:input_nle}
\end{table}

\subsection{Further Analysis}
\label{sec:suppl}

\paragraph{Why is \casethree producing the most accurate results?} Our study demonstrates that \casethree surpasses both \casetwo and \casefour in accuracy. However, the reasons behind this superior performance are not yet understood. Therefore, this section focuses on systematically analyzing the efficacy of \casethree.

We first dissect the effectiveness of \casethree over \casetwo. As shown in \tabref{tab:input_nle}, NLEs from \casetwo are mere verbatim copies of inputs rather than insightful explanations. To substantiate this, we calculate the ROUGE-L scores between the NAN test set and the corresponding NLEs from \casetwo and \casethree as a means of similarity measurement. As depicted in \figref{fig:rouge_nle}, NLEs from \casetwo often replicate the given premise and hypothesis, resulting in high ROUGE-L scores. Instead, \casethree can produce meaningful NLEs, demonstrating lower similarity to the test instances.

\begin{figure}
    \centering
    \includegraphics[width=0.98\linewidth]{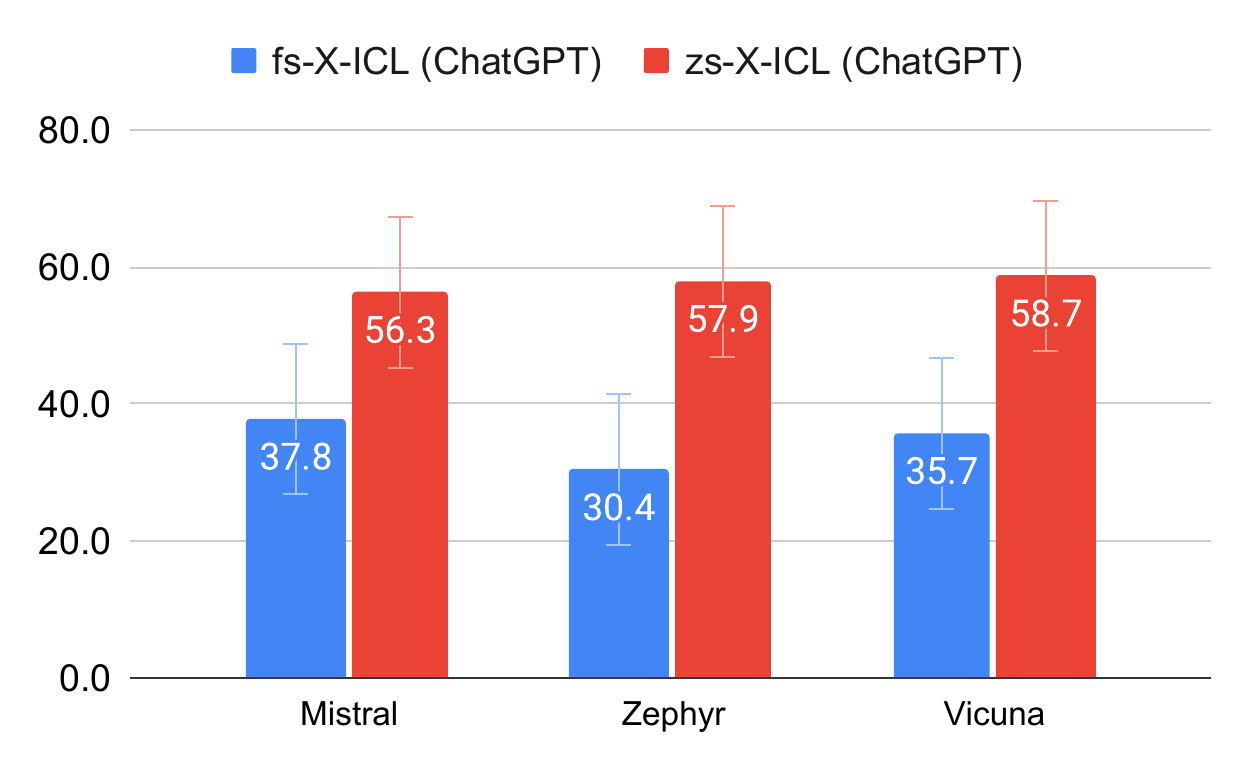}
    \caption{Average length (\#words) of NLEs generated by \casethree and \casefour.}
    \label{fig:nle_len}
\end{figure}

\begin{table}[]
    \centering
    \scalebox{0.83}{
    \begin{tabular}{lccc}
    \toprule
    \textbf{Methods}&\textbf{Mistral} & \textbf{Zephyr} & \textbf{Vicuna}\\
    \midrule
      \casetwo  & 53.5 & 59.3 & 59.8\\
      \casefour & 46.4 & 58.1 & 58.8\\
      \caseeight & 56.2 & 62.3 & 63.4\\
      \casethree & 57.1 & 65.5 & 62.1\\
    \bottomrule
    \end{tabular}}
    \caption{Average accuracy of \casetwo, \casefour, \caseeight and \casethree among all test sets.}
    \label{tab:short-nle}
\end{table}

After analyzing the NLEs from \casefour, we attribute the inefficiency to verbose NLEs. Specifically, \figref{fig:nle_len} shows that \casefour produces longer NLEs than \casethree. As a result, we observe inconsistency within the NLEs, leading to incorrect predictions. As a remedy, we prompt ChatGPT to generate shorter NLEs in the zero-shot setting, denoted as \caseeight. Compared to \casefour, the NLEs generated by \caseeight are reduced to an average of 27 tokens. Consequently, with the help of the concise NLEs, we can improve the accuracy significantly and even surpass the \casetwo as shown in \tabref{tab:short-nle}.

\section{Summary and Outlook}
We introduced a simple yet effective method called \casethree, leveraging human-written NLEs to generate synthetic NLEs by prompting ChatGPT. \casethree significantly boosts accuracy across various adversarial datasets and five LLMs, compared to standard in-context learning and \xicl using human-written NLEs. {Additionally, our analysis revealed that data selection methodologies may exhibit overfitting within the in-distribution dataset, thus potentially failing to extend to unseen or adversarial datasets. In contrast, our approach employing NLEs has shown consistent performance in both in-distribution and adversarial contexts.}
Our work paves the way for more robust performance and enhanced explainability capabilities of LLMs.

\section*{Limitations}
One limitation of \xicl might be the observed lack of fidelity in the NLEs generated by LLMs, despite their capability to provide accurate answers. These NLEs may sometimes include unfaithful or hallucinated information, which if relied upon by users for model trust, can lead to severe implications. Testing and enhancing the faithfulness of NLEs is a challenging open question \citep{pepa}. In this work, we show that \xicl improves robustness, but we do not advocate using the generated NLEs as faithful explanations without further testing. 
Second, our approach exhibited promising results when tested against adversarial datasets in two notable NLP tasks: natural language inference and paraphrasing identification. However, further research is required to examine the performance of LLMs and their generalizability across diverse NLP tasks in the context of adversarial examples. 

\section*{Acknowledgements}
Xuanli He was supported by an industry grant from Cisco. Oana-Maria Camburu was supported by a Leverhulme Early Career Fellowship. 
Pasquale Minervini was partially funded by the European Union’s Horizon 2020 research and innovation programme under grant agreement no. 875160, ELIAI (The Edinburgh Laboratory for Integrated Artificial Intelligence) EPSRC (grant no. EP/W002876/1), an industry grant from Cisco, and a donation from Accenture LLP; and is grateful to NVIDIA for the GPU donations.
This work was supported by the Edinburgh International Data Facility (EIDF) and the Data-Driven Innovation Programme at the University of Edinburgh.


%

%

\bibliography{anthology,custom}

\begin{thebibliography}{65}
\expandafter\ifx\csname natexlab\endcsname\relax\def\natexlab#1{#1}\fi

\bibitem[{Alvarez-Melis and
  Jaakkola(2017)}]{alvarez-melis-jaakkola-2017-causal}
David Alvarez-Melis and Tommi Jaakkola. 2017.
\newblock \href {https://doi.org/10.18653/v1/D17-1042} {A causal framework for
  explaining the predictions of black-box sequence-to-sequence models}.
\newblock In \emph{Proceedings of the 2017 Conference on Empirical Methods in
  Natural Language Processing}, pages 412--421, Copenhagen, Denmark.
  Association for Computational Linguistics.

\bibitem[{Atanasova et~al.(2023)Atanasova, Camburu, Lioma, Lukasiewicz,
  Simonsen, and Augenstein}]{pepa}
Pepa Atanasova, Oana-Maria Camburu, Christina Lioma, Thomas Lukasiewicz,
  Jakob~Grue Simonsen, and Isabelle Augenstein. 2023.
\newblock {Faithfulness Tests for Natural Language Explanations}.
\newblock In \emph{ACL}.

\bibitem[{Bang et~al.(2023)Bang, Cahyawijaya, Lee, Dai, Su, Wilie, Lovenia, Ji,
  Yu, Chung, Do, Xu, and Fung}]{DBLP:journals/corr/abs-2302-04023}
Yejin Bang, Samuel Cahyawijaya, Nayeon Lee, Wenliang Dai, Dan Su, Bryan Wilie,
  Holy Lovenia, Ziwei Ji, Tiezheng Yu, Willy Chung, Quyet~V. Do, Yan Xu, and
  Pascale Fung. 2023.
\newblock A multitask, multilingual, multimodal evaluation of chatgpt on
  reasoning, hallucination, and interactivity.
\newblock \emph{CoRR}, abs/2302.04023.

\bibitem[{Bowman et~al.(2015)Bowman, Angeli, Potts, and
  Manning}]{bowman-etal-2015-large}
Samuel~R. Bowman, Gabor Angeli, Christopher Potts, and Christopher~D. Manning.
  2015.
\newblock A large annotated corpus for learning natural language inference.
\newblock In \emph{{EMNLP}}, pages 632--642. The Association for Computational
  Linguistics.

\bibitem[{Brown et~al.(2020)Brown, Mann, Ryder, Subbiah, Kaplan, Dhariwal,
  Neelakantan, Shyam, Sastry, Askell, Agarwal, Herbert-Voss, Krueger, Henighan,
  Child, Ramesh, Ziegler, Wu, Winter, Hesse, Chen, Sigler, Litwin, Gray, Chess,
  Clark, Berner, McCandlish, Radford, Sutskever, and
  Amodei}]{NEURIPS2020_1457c0d6}
Tom Brown, Benjamin Mann, Nick Ryder, Melanie Subbiah, Jared~D Kaplan, Prafulla
  Dhariwal, Arvind Neelakantan, Pranav Shyam, Girish Sastry, Amanda Askell,
  Sandhini Agarwal, Ariel Herbert-Voss, Gretchen Krueger, Tom Henighan, Rewon
  Child, Aditya Ramesh, Daniel Ziegler, Jeffrey Wu, Clemens Winter, Chris
  Hesse, Mark Chen, Eric Sigler, Mateusz Litwin, Scott Gray, Benjamin Chess,
  Jack Clark, Christopher Berner, Sam McCandlish, Alec Radford, Ilya Sutskever,
  and Dario Amodei. 2020.
\newblock \href
  {https://proceedings.neurips.cc/paper_files/paper/2020/file/1457c0d6bfcb4967418bfb8ac142f64a-Paper.pdf}
  {Language models are few-shot learners}.
\newblock In \emph{Advances in Neural Information Processing Systems},
  volume~33, pages 1877--1901. Curran Associates, Inc.

\bibitem[{Camburu et~al.(2018)Camburu, Rockt{\"a}schel, Lukasiewicz, and
  Blunsom}]{camburu2018snli}
Oana-Maria Camburu, Tim Rockt{\"a}schel, Thomas Lukasiewicz, and Phil Blunsom.
  2018.
\newblock e-snli: Natural language inference with natural language
  explanations.
\newblock \emph{Advances in Neural Information Processing Systems}, 31.

\bibitem[{Carlini et~al.(2023)Carlini, Ippolito, Jagielski, Lee, Tramer, and
  Zhang}]{carlini2023quantifying}
Nicholas Carlini, Daphne Ippolito, Matthew Jagielski, Katherine Lee, Florian
  Tramer, and Chiyuan Zhang. 2023.
\newblock \href {https://openreview.net/forum?id=TatRHT_1cK} {Quantifying
  memorization across neural language models}.
\newblock In \emph{The Eleventh International Conference on Learning
  Representations}.

\bibitem[{Chen et~al.(2022)Chen, He, Narasimhan, and Chen}]{chen2022can}
Howard Chen, Jacqueline He, Karthik Narasimhan, and Danqi Chen. 2022.
\newblock Can rationalization improve robustness?
\newblock In \emph{Proceedings of the 2022 Conference of the North American
  Chapter of the Association for Computational Linguistics: Human Language
  Technologies}, pages 3792--3805.

\bibitem[{Chiang et~al.(2023)Chiang, Li, Lin, Sheng, Wu, Zhang, Zheng, Zhuang,
  Zhuang, Gonzalez, Stoica, and Xing}]{vicuna2023}
Wei-Lin Chiang, Zhuohan Li, Zi~Lin, Ying Sheng, Zhanghao Wu, Hao Zhang, Lianmin
  Zheng, Siyuan Zhuang, Yonghao Zhuang, Joseph~E. Gonzalez, Ion Stoica, and
  Eric~P. Xing. 2023.
\newblock \href {https://lmsys.org/blog/2023-03-30-vicuna/} {Vicuna: An
  open-source chatbot impressing gpt-4 with 90\%* chatgpt quality}.

\bibitem[{Clark et~al.(2019)Clark, Yatskar, and
  Zettlemoyer}]{clark-etal-2019-dont}
Christopher Clark, Mark Yatskar, and Luke Zettlemoyer. 2019.
\newblock \href {https://doi.org/10.18653/v1/D19-1418} {Don{'}t take the easy
  way out: Ensemble based methods for avoiding known dataset biases}.
\newblock In \emph{Proceedings of the 2019 Conference on Empirical Methods in
  Natural Language Processing and the 9th International Joint Conference on
  Natural Language Processing (EMNLP-IJCNLP)}, pages 4069--4082, Hong Kong,
  China. Association for Computational Linguistics.

\bibitem[{Guo et~al.(2023)Guo, Zhang, Wang, Jiang, Nie, Ding, Yue, and
  Wu}]{DBLP:journals/corr/abs-2301-07597}
Biyang Guo, Xin Zhang, Ziyuan Wang, Minqi Jiang, Jinran Nie, Yuxuan Ding,
  Jianwei Yue, and Yupeng Wu. 2023.
\newblock How close is chatgpt to human experts? comparison corpus, evaluation,
  and detection.
\newblock \emph{CoRR}, abs/2301.07597.

\bibitem[{Gupta et~al.(2023)Gupta, Singh, and Gardner}]{gupta2023coverage}
Shivanshu Gupta, Sameer Singh, and Matt Gardner. 2023.
\newblock Coverage-based example selection for in-context learning.
\newblock \emph{arXiv preprint arXiv:2305.14907}.

\bibitem[{Gururangan et~al.(2018)Gururangan, Swayamdipta, Levy, Schwartz,
  Bowman, and Smith}]{gururangan-etal-2018-annotation}
Suchin Gururangan, Swabha Swayamdipta, Omer Levy, Roy Schwartz, Samuel Bowman,
  and Noah~A. Smith. 2018.
\newblock \href {https://doi.org/10.18653/v1/N18-2017} {Annotation artifacts in
  natural language inference data}.
\newblock In \emph{Proceedings of the 2018 Conference of the North {A}merican
  Chapter of the Association for Computational Linguistics: Human Language
  Technologies, Volume 2 (Short Papers)}, pages 107--112, New Orleans,
  Louisiana. Association for Computational Linguistics.

\bibitem[{Hase et~al.(2020)Hase, Zhang, Xie, and
  Bansal}]{hase-etal-2020-leakage}
Peter Hase, Shiyue Zhang, Harry Xie, and Mohit Bansal. 2020.
\newblock \href {https://doi.org/10.18653/v1/2020.findings-emnlp.390}
  {Leakage-adjusted simulatability: Can models generate non-trivial
  explanations of their behavior in natural language?}
\newblock In \emph{Findings of the Association for Computational Linguistics:
  EMNLP 2020}, pages 4351--4367, Online. Association for Computational
  Linguistics.

\bibitem[{He et~al.(2019)He, Zha, and Wang}]{he-etal-2019-unlearn}
He~He, Sheng Zha, and Haohan Wang. 2019.
\newblock \href {https://doi.org/10.18653/v1/D19-6115} {Unlearn dataset bias in
  natural language inference by fitting the residual}.
\newblock In \emph{Proceedings of the 2nd Workshop on Deep Learning Approaches
  for Low-Resource NLP (DeepLo 2019)}, pages 132--142, Hong Kong, China.
  Association for Computational Linguistics.

\bibitem[{Hendricks et~al.(2018)Hendricks, Hu, Darrell, and
  Akata}]{Hendricks_2018_ECCV}
Lisa~Anne Hendricks, Ronghang Hu, Trevor Darrell, and Zeynep Akata. 2018.
\newblock Grounding visual explanations.
\newblock In \emph{Proceedings of the European Conference on Computer Vision
  (ECCV)}.

\bibitem[{Jiang et~al.(2023)Jiang, Sablayrolles, Mensch, Bamford, Chaplot,
  Casas, Bressand, Lengyel, Lample, Saulnier et~al.}]{jiang2023mistral}
Albert~Q Jiang, Alexandre Sablayrolles, Arthur Mensch, Chris Bamford,
  Devendra~Singh Chaplot, Diego de~las Casas, Florian Bressand, Gianna Lengyel,
  Guillaume Lample, Lucile Saulnier, et~al. 2023.
\newblock Mistral 7b.
\newblock \emph{arXiv preprint arXiv:2310.06825}.

\bibitem[{Karimi~Mahabadi et~al.(2020)Karimi~Mahabadi, Belinkov, and
  Henderson}]{karimi-mahabadi-etal-2020-end}
Rabeeh Karimi~Mahabadi, Yonatan Belinkov, and James Henderson. 2020.
\newblock \href {https://doi.org/10.18653/v1/2020.acl-main.769} {End-to-end
  bias mitigation by modelling biases in corpora}.
\newblock In \emph{Proceedings of the 58th Annual Meeting of the Association
  for Computational Linguistics}, pages 8706--8716, Online. Association for
  Computational Linguistics.

\bibitem[{Kavumba et~al.(2023)Kavumba, Brassard, Heinzerling, and
  Inui}]{kavumba-etal-2023-prompting}
Pride Kavumba, Ana Brassard, Benjamin Heinzerling, and Kentaro Inui. 2023.
\newblock \href {https://doi.org/10.18653/v1/2023.findings-eacl.162} {Prompting
  for explanations improves adversarial {NLI}. is this true? {Yes} it is {true}
  because {it weakens superficial cues}}.
\newblock In \emph{Findings of the Association for Computational Linguistics:
  EACL 2023}, pages 2165--2180, Dubrovnik, Croatia. Association for
  Computational Linguistics.

\bibitem[{Kayser et~al.(2021)Kayser, Camburu, Salewski, Emde, Do, Akata, and
  Lukasiewicz}]{kayser2021vil}
Maxime Kayser, Oana-Maria Camburu, Leonard Salewski, Cornelius Emde, Virginie
  Do, Zeynep Akata, and Thomas Lukasiewicz. 2021.
\newblock {e-ViL: A} dataset and benchmark for natural language explanations in
  vision-language tasks.
\newblock In \emph{Proceedings of the IEEE/CVF International Conference on
  Computer Vision}, pages 1244--1254.

\bibitem[{Kayser et~al.(2022)Kayser, Emde, Camburu, Parsons, Papiez, and
  Lukasiewicz}]{mimicnle}
Maxime Kayser, Cornelius Emde, Oana-Maria Camburu, Guy Parsons, Bartlomiej
  Papiez, and Thomas Lukasiewicz. 2022.
\newblock Explaining chest x-ray pathologies in natural language.
\newblock In \emph{Medical Image Computing and Computer Assisted Intervention
  -- MICCAI 2022}, pages 701--713, Cham. Springer Nature Switzerland.

\bibitem[{Kim et~al.(2018)Kim, Rohrbach, Darrell, Canny, and Akata}]{cars}
Jinkyu Kim, Anna Rohrbach, Trevor Darrell, John~F. Canny, and Zeynep Akata.
  2018.
\newblock \href {http://arxiv.org/abs/1807.11546} {Textual explanations for
  self-driving vehicles}.
\newblock \emph{CoRR}, abs/1807.11546.

\bibitem[{Korakakis and Vlachos(2023)}]{korakakis-vlachos-2023-improving}
Michalis Korakakis and Andreas Vlachos. 2023.
\newblock \href {https://doi.org/10.18653/v1/2023.acl-long.801} {Improving the
  robustness of {NLI} models with minimax training}.
\newblock In \emph{Proceedings of the 61st Annual Meeting of the Association
  for Computational Linguistics (Volume 1: Long Papers)}, pages 14322--14339,
  Toronto, Canada. Association for Computational Linguistics.

\bibitem[{Levy et~al.(2023)Levy, Bogin, and Berant}]{levy-etal-2023-diverse}
Itay Levy, Ben Bogin, and Jonathan Berant. 2023.
\newblock \href {https://doi.org/10.18653/v1/2023.acl-long.78} {Diverse
  demonstrations improve in-context compositional generalization}.
\newblock In \emph{Proceedings of the 61st Annual Meeting of the Association
  for Computational Linguistics (Volume 1: Long Papers)}, pages 1401--1422,
  Toronto, Canada. Association for Computational Linguistics.

\bibitem[{Li et~al.(2022)Li, Chen, Shen, Chen, Zhang, Li, Wang, Qian, Peng, Mao
  et~al.}]{li2022explanations}
Shiyang Li, Jianshu Chen, Yelong Shen, Zhiyu Chen, Xinlu Zhang, Zekun Li, Hong
  Wang, Jing Qian, Baolin Peng, Yi~Mao, et~al. 2022.
\newblock Explanations from large language models make small reasoners better.
\newblock \emph{arXiv preprint arXiv:2210.06726}.

\bibitem[{Liu et~al.(2022)Liu, Shen, Zhang, Dolan, Carin, and
  Chen}]{liu-etal-2022-makes}
Jiachang Liu, Dinghan Shen, Yizhe Zhang, Bill Dolan, Lawrence Carin, and Weizhu
  Chen. 2022.
\newblock \href {https://doi.org/10.18653/v1/2022.deelio-1.10} {What makes good
  in-context examples for {GPT}-3?}
\newblock In \emph{Proceedings of Deep Learning Inside Out (DeeLIO 2022): The
  3rd Workshop on Knowledge Extraction and Integration for Deep Learning
  Architectures}, pages 100--114, Dublin, Ireland and Online. Association for
  Computational Linguistics.

\bibitem[{Liu et~al.(2020{\natexlab{a}})Liu, Xin, Chang, and
  Sui}]{liu-etal-2020-hyponli}
Tianyu Liu, Zheng Xin, Baobao Chang, and Zhifang Sui. 2020{\natexlab{a}}.
\newblock \href {https://aclanthology.org/2020.lrec-1.846} {{H}ypo{NLI}:
  Exploring the artificial patterns of hypothesis-only bias in natural language
  inference}.
\newblock In \emph{Proceedings of the Twelfth Language Resources and Evaluation
  Conference}, pages 6852--6860, Marseille, France. European Language Resources
  Association.

\bibitem[{Liu et~al.(2020{\natexlab{b}})Liu, Xin, Ding, Chang, and
  Sui}]{liu-etal-2020-empirical}
Tianyu Liu, Zheng Xin, Xiaoan Ding, Baobao Chang, and Zhifang Sui.
  2020{\natexlab{b}}.
\newblock \href {https://doi.org/10.18653/v1/2020.conll-1.48} {An empirical
  study on model-agnostic debiasing strategies for robust natural language
  inference}.
\newblock In \emph{Proceedings of the 24th Conference on Computational Natural
  Language Learning}, pages 596--608, Online. Association for Computational
  Linguistics.

\bibitem[{Lu et~al.(2022)Lu, Bartolo, Moore, Riedel, and
  Stenetorp}]{lu-etal-2022-fantastically}
Yao Lu, Max Bartolo, Alastair Moore, Sebastian Riedel, and Pontus Stenetorp.
  2022.
\newblock \href {https://doi.org/10.18653/v1/2022.acl-long.556} {Fantastically
  ordered prompts and where to find them: Overcoming few-shot prompt order
  sensitivity}.
\newblock In \emph{Proceedings of the 60th Annual Meeting of the Association
  for Computational Linguistics (Volume 1: Long Papers)}, pages 8086--8098,
  Dublin, Ireland. Association for Computational Linguistics.

\bibitem[{Ludan et~al.(2023)Ludan, Meng, Nguyen, Shah, Lyu, Apidianaki, and
  Callison-Burch}]{ludan-etal-2023-explanation}
Josh~Magnus Ludan, Yixuan Meng, Tai Nguyen, Saurabh Shah, Qing Lyu, Marianna
  Apidianaki, and Chris Callison-Burch. 2023.
\newblock \href {https://doi.org/10.18653/v1/2023.acl-long.242}
  {Explanation-based finetuning makes models more robust to spurious cues}.
\newblock In \emph{Proceedings of the 61st Annual Meeting of the Association
  for Computational Linguistics (Volume 1: Long Papers)}, pages 4420--4441,
  Toronto, Canada. Association for Computational Linguistics.

\bibitem[{Majumder et~al.(2022)Majumder, Camburu, Lukasiewicz, and
  Mcauley}]{rexc}
Bodhisattwa~Prasad Majumder, Oana-Maria Camburu, Thomas Lukasiewicz, and Julian
  Mcauley. 2022.
\newblock Knowledge-grounded self-rationalization via extractive and natural
  language explanations.
\newblock In \emph{Proceedings of the 39th International Conference on Machine
  Learning}, volume 162 of \emph{Proceedings of Machine Learning Research},
  pages 14786--14801. PMLR.

\bibitem[{McCoy et~al.(2019)McCoy, Pavlick, and Linzen}]{mccoy-etal-2019-right}
Tom McCoy, Ellie Pavlick, and Tal Linzen. 2019.
\newblock \href {https://doi.org/10.18653/v1/P19-1334} {Right for the wrong
  reasons: Diagnosing syntactic heuristics in natural language inference}.
\newblock In \emph{Proceedings of the 57th Annual Meeting of the Association
  for Computational Linguistics}, pages 3428--3448, Florence, Italy.
  Association for Computational Linguistics.

\bibitem[{Minervini and Riedel(2018)}]{minervini-riedel-2018-adversarially}
Pasquale Minervini and Sebastian Riedel. 2018.
\newblock \href {https://doi.org/10.18653/v1/K18-1007} {Adversarially
  regularising neural {NLI} models to integrate logical background knowledge}.
\newblock In \emph{Proceedings of the 22nd Conference on Computational Natural
  Language Learning}, pages 65--74, Brussels, Belgium. Association for
  Computational Linguistics.

\bibitem[{Naik et~al.(2018)Naik, Ravichander, Sadeh, Rose, and
  Neubig}]{naik-etal-2018-stress}
Aakanksha Naik, Abhilasha Ravichander, Norman Sadeh, Carolyn Rose, and Graham
  Neubig. 2018.
\newblock \href {https://aclanthology.org/C18-1198} {Stress test evaluation for
  natural language inference}.
\newblock In \emph{Proceedings of the 27th International Conference on
  Computational Linguistics}, pages 2340--2353, Santa Fe, New Mexico, USA.
  Association for Computational Linguistics.

\bibitem[{Narang et~al.(2020)Narang, Raffel, Lee, Roberts, Fiedel, and
  Malkan}]{narang2020wt5}
Sharan Narang, Colin Raffel, Katherine Lee, Adam Roberts, Noah Fiedel, and
  Karishma Malkan. 2020.
\newblock Wt5?! training text-to-text models to explain their predictions.
\newblock \emph{arXiv preprint arXiv:2004.14546}.

\bibitem[{Nie et~al.(2019)Nie, Wang, and Bansal}]{nie2019analyzing}
Yixin Nie, Yicheng Wang, and Mohit Bansal. 2019.
\newblock Analyzing compositionality-sensitivity of nli models.
\newblock In \emph{Proceedings of the AAAI Conference on Artificial
  Intelligence}, volume~33, pages 6867--6874.

\bibitem[{Nie et~al.(2020)Nie, Williams, Dinan, Bansal, Weston, and
  Kiela}]{nie-etal-2020-adversarial}
Yixin Nie, Adina Williams, Emily Dinan, Mohit Bansal, Jason Weston, and Douwe
  Kiela. 2020.
\newblock Adversarial {NLI:} {A} new benchmark for natural language
  understanding.
\newblock In \emph{{ACL}}, pages 4885--4901. Association for Computational
  Linguistics.

\bibitem[{Rae et~al.(2021)Rae, Borgeaud, Cai, Millican, Hoffmann, Song,
  Aslanides, Henderson, Ring, Young et~al.}]{rae2021scaling}
Jack~W Rae, Sebastian Borgeaud, Trevor Cai, Katie Millican, Jordan Hoffmann,
  Francis Song, John Aslanides, Sarah Henderson, Roman Ring, Susannah Young,
  et~al. 2021.
\newblock Scaling language models: Methods, analysis \& insights from training
  gopher.
\newblock \emph{arXiv preprint arXiv:2112.11446}.

\bibitem[{Rajani et~al.(2019)Rajani, McCann, Xiong, and
  Socher}]{rajani-etal-2019-explain}
Nazneen~Fatema Rajani, Bryan McCann, Caiming Xiong, and Richard Socher. 2019.
\newblock \href {https://doi.org/10.18653/v1/P19-1487} {Explain yourself!
  leveraging language models for commonsense reasoning}.
\newblock In \emph{Proceedings of the 57th Annual Meeting of the Association
  for Computational Linguistics}, pages 4932--4942, Florence, Italy.
  Association for Computational Linguistics.

\bibitem[{Reimers and Gurevych(2019)}]{reimers-gurevych-2019-sentence}
Nils Reimers and Iryna Gurevych. 2019.
\newblock \href {https://doi.org/10.18653/v1/D19-1410} {Sentence-{BERT}:
  Sentence embeddings using {S}iamese {BERT}-networks}.
\newblock In \emph{Proceedings of the 2019 Conference on Empirical Methods in
  Natural Language Processing and the 9th International Joint Conference on
  Natural Language Processing (EMNLP-IJCNLP)}, pages 3982--3992, Hong Kong,
  China. Association for Computational Linguistics.

\bibitem[{Ribeiro et~al.(2016)Ribeiro, Singh, and
  Guestrin}]{10.1145/2939672.2939778}
Marco~Tulio Ribeiro, Sameer Singh, and Carlos Guestrin. 2016.
\newblock \href {https://doi.org/10.1145/2939672.2939778} {"why should i trust
  you?": Explaining the predictions of any classifier}.
\newblock In \emph{Proceedings of the 22nd ACM SIGKDD International Conference
  on Knowledge Discovery and Data Mining}, KDD '16, page 1135–1144, New York,
  NY, USA. Association for Computing Machinery.

\bibitem[{Sainz et~al.(2023)Sainz, Campos, García-Ferrero, Etxaniz, and
  Agirre}]{sainz2023}
Oscar Sainz, Jon~Ander Campos, Iker García-Ferrero, Julen Etxaniz, and Eneko
  Agirre. 2023.
\newblock \href {https://hitz-zentroa.github.io/lm-contamination/blog/} {Did
  chatgpt cheat on your test?}

\bibitem[{Scott(1962)}]{d69ed548-5302-3d45-87a7-4fea0bc35a5e}
William~A. Scott. 1962.
\newblock \href {http://www.jstor.org/stable/2785779} {Cognitive complexity and
  cognitive flexibility}.
\newblock \emph{Sociometry}, 25(4):405--414.

\bibitem[{Serrano and Smith(2019)}]{serrano-smith-2019-attention}
Sofia Serrano and Noah~A. Smith. 2019.
\newblock \href {https://doi.org/10.18653/v1/P19-1282} {Is attention
  interpretable?}
\newblock In \emph{Proceedings of the 57th Annual Meeting of the Association
  for Computational Linguistics}, pages 2931--2951, Florence, Italy.
  Association for Computational Linguistics.

\bibitem[{{Sparck Jones} et~al.(2000){Sparck Jones}, Walker, and
  Robertson}]{SPARCKJONES2000779}
K.~{Sparck Jones}, S.~Walker, and S.E. Robertson. 2000.
\newblock \href {https://doi.org/https://doi.org/10.1016/S0306-4573(00)00015-7}
  {A probabilistic model of information retrieval: development and comparative
  experiments: Part 1}.
\newblock \emph{Information Processing and Management}, 36(6):779--808.

\bibitem[{Srivastava et~al.(2022)Srivastava, Rastogi, Rao, Shoeb, Abid, Fisch,
  Brown, Santoro, Gupta, Garriga-Alonso et~al.}]{srivastava2022beyond}
Aarohi Srivastava, Abhinav Rastogi, Abhishek Rao, Abu Awal~Md Shoeb, Abubakar
  Abid, Adam Fisch, Adam~R Brown, Adam Santoro, Aditya Gupta, Adri{\`a}
  Garriga-Alonso, et~al. 2022.
\newblock Beyond the imitation game: Quantifying and extrapolating the
  capabilities of language models.
\newblock \emph{arXiv preprint arXiv:2206.04615}.

\bibitem[{Stacey et~al.(2022)Stacey, Belinkov, and Rei}]{stacey2022supervising}
Joe Stacey, Yonatan Belinkov, and Marek Rei. 2022.
\newblock Supervising model attention with human explanations for robust
  natural language inference.
\newblock In \emph{Proceedings of the AAAI Conference on Artificial
  Intelligence}, volume~36, pages 11349--11357.

\bibitem[{Touvron et~al.(2023)Touvron, Martin, Stone, Albert, Almahairi,
  Babaei, Bashlykov, Batra, Bhargava, Bhosale et~al.}]{touvron2023llama}
Hugo Touvron, Louis Martin, Kevin Stone, Peter Albert, Amjad Almahairi, Yasmine
  Babaei, Nikolay Bashlykov, Soumya Batra, Prajjwal Bhargava, Shruti Bhosale,
  et~al. 2023.
\newblock Llama 2: Open foundation and fine-tuned chat models.
\newblock \emph{arXiv preprint arXiv:2307.09288}.

\bibitem[{Truong et~al.(2022)Truong, Otmakhova, Baldwin, Cohn, Lau, and
  Verspoor}]{truong-etal-2022-another}
Thinh~Hung Truong, Yulia Otmakhova, Timothy Baldwin, Trevor Cohn, Jey~Han Lau,
  and Karin Verspoor. 2022.
\newblock \href {https://aclanthology.org/2022.aacl-main.65} {Not another
  negation benchmark: The {N}a{N}-{NLI} test suite for sub-clausal negation}.
\newblock In \emph{Proceedings of the 2nd Conference of the Asia-Pacific
  Chapter of the Association for Computational Linguistics and the 12th
  International Joint Conference on Natural Language Processing (Volume 1: Long
  Papers)}, pages 883--894, Online only. Association for Computational
  Linguistics.

\bibitem[{Tunstall et~al.(2023)Tunstall, Beeching, Lambert, Rajani, Rasul,
  Belkada, Huang, von Werra, Fourrier, Habib et~al.}]{tunstall2023zephyr}
Lewis Tunstall, Edward Beeching, Nathan Lambert, Nazneen Rajani, Kashif Rasul,
  Younes Belkada, Shengyi Huang, Leandro von Werra, Cl{\'e}mentine Fourrier,
  Nathan Habib, et~al. 2023.
\newblock Zephyr: Direct distillation of lm alignment.
\newblock \emph{arXiv preprint arXiv:2310.16944}.

\bibitem[{Wang et~al.(2018)Wang, Singh, Michael, Hill, Levy, and
  Bowman}]{wang-etal-2018-glue}
Alex Wang, Amanpreet Singh, Julian Michael, Felix Hill, Omer Levy, and Samuel
  Bowman. 2018.
\newblock \href {https://doi.org/10.18653/v1/W18-5446} {{GLUE}: A multi-task
  benchmark and analysis platform for natural language understanding}.
\newblock In \emph{Proceedings of the 2018 {EMNLP} Workshop {B}lackbox{NLP}:
  Analyzing and Interpreting Neural Networks for {NLP}}, pages 353--355,
  Brussels, Belgium. Association for Computational Linguistics.

\bibitem[{Wang et~al.(2023{\natexlab{a}})Wang, Liu, Park, Chen, and
  Xiao}]{wang2023adversarial}
Jiongxiao Wang, Zichen Liu, Keun~Hee Park, Muhao Chen, and Chaowei Xiao.
  2023{\natexlab{a}}.
\newblock Adversarial demonstration attacks on large language models.
\newblock \emph{arXiv preprint arXiv:2305.14950}.

\bibitem[{Wang et~al.(2023{\natexlab{b}})Wang, Wei, Schuurmans, Le, Chi,
  Narang, Chowdhery, and Zhou}]{wang2023selfconsistency}
Xuezhi Wang, Jason Wei, Dale Schuurmans, Quoc~V Le, Ed~H. Chi, Sharan Narang,
  Aakanksha Chowdhery, and Denny Zhou. 2023{\natexlab{b}}.
\newblock \href {https://openreview.net/forum?id=1PL1NIMMrw} {Self-consistency
  improves chain of thought reasoning in language models}.
\newblock In \emph{The Eleventh International Conference on Learning
  Representations}.

\bibitem[{Wei et~al.(2022{\natexlab{a}})Wei, Tay, Bommasani, Raffel, Zoph,
  Borgeaud, Yogatama, Bosma, Zhou, Metzler, Chi, Hashimoto, Vinyals, Liang,
  Dean, and Fedus}]{wei2022emergent}
Jason Wei, Yi~Tay, Rishi Bommasani, Colin Raffel, Barret Zoph, Sebastian
  Borgeaud, Dani Yogatama, Maarten Bosma, Denny Zhou, Donald Metzler, Ed~H.
  Chi, Tatsunori Hashimoto, Oriol Vinyals, Percy Liang, Jeff Dean, and William
  Fedus. 2022{\natexlab{a}}.
\newblock \href {https://openreview.net/forum?id=yzkSU5zdwD} {Emergent
  abilities of large language models}.
\newblock \emph{Transactions on Machine Learning Research}.
\newblock Survey Certification.

\bibitem[{Wei et~al.(2022{\natexlab{b}})Wei, Wang, Schuurmans, Bosma, Ichter,
  Xia, Chi, Le, and Zhou}]{wei2022chain}
Jason Wei, Xuezhi Wang, Dale Schuurmans, Maarten Bosma, Brian Ichter, Fei Xia,
  Ed~H. Chi, Quoc~V. Le, and Denny Zhou. 2022{\natexlab{b}}.
\newblock Chain-of-thought prompting elicits reasoning in large language
  models.
\newblock In \emph{NeurIPS}.

\bibitem[{Wiegreffe and Marasovic(2021)}]{teachmetoexplain}
Sarah Wiegreffe and Ana Marasovic. 2021.
\newblock Teach me to explain: A review of datasets for explainable natural
  language processing.
\newblock \emph{35th Conference on Neural Information Processing Systems
  (NeurIPS) Track on Datasets and Benchmarks}.

\bibitem[{Williams et~al.(2018)Williams, Nangia, and
  Bowman}]{williams-etal-2018-broad}
Adina Williams, Nikita Nangia, and Samuel Bowman. 2018.
\newblock \href {https://doi.org/10.18653/v1/N18-1101} {A broad-coverage
  challenge corpus for sentence understanding through inference}.
\newblock In \emph{Proceedings of the 2018 Conference of the North {A}merican
  Chapter of the Association for Computational Linguistics: Human Language
  Technologies, Volume 1 (Long Papers)}, pages 1112--1122, New Orleans,
  Louisiana. Association for Computational Linguistics.

\bibitem[{Wu et~al.(2021)Wu, Ribeiro, Heer, and Weld}]{wu-etal-2021-polyjuice}
Tongshuang Wu, Marco~Tulio Ribeiro, Jeffrey Heer, and Daniel Weld. 2021.
\newblock \href {https://doi.org/10.18653/v1/2021.acl-long.523} {Polyjuice:
  Generating counterfactuals for explaining, evaluating, and improving models}.
\newblock In \emph{Proceedings of the 59th Annual Meeting of the Association
  for Computational Linguistics and the 11th International Joint Conference on
  Natural Language Processing (Volume 1: Long Papers)}, pages 6707--6723,
  Online. Association for Computational Linguistics.

\bibitem[{Wu et~al.(2022)Wu, Gardner, Stenetorp, and
  Dasigi}]{wu-etal-2022-generating}
Yuxiang Wu, Matt Gardner, Pontus Stenetorp, and Pradeep Dasigi. 2022.
\newblock \href {https://doi.org/10.18653/v1/2022.acl-long.190} {Generating
  data to mitigate spurious correlations in natural language inference
  datasets}.
\newblock In \emph{Proceedings of the 60th Annual Meeting of the Association
  for Computational Linguistics (Volume 1: Long Papers)}, pages 2660--2676,
  Dublin, Ireland. Association for Computational Linguistics.

\bibitem[{Yaghoobzadeh et~al.(2021)Yaghoobzadeh, Mehri, Tachet~des Combes,
  Hazen, and Sordoni}]{yaghoobzadeh-etal-2021-increasing}
Yadollah Yaghoobzadeh, Soroush Mehri, Remi Tachet~des Combes, T.~J. Hazen, and
  Alessandro Sordoni. 2021.
\newblock \href {https://doi.org/10.18653/v1/2021.eacl-main.291} {Increasing
  robustness to spurious correlations using forgettable examples}.
\newblock In \emph{Proceedings of the 16th Conference of the European Chapter
  of the Association for Computational Linguistics: Main Volume}, pages
  3319--3332, Online. Association for Computational Linguistics.

\bibitem[{Ye et~al.(2023)Ye, Wu, Feng, Yu, and Kong}]{10.5555/3618408.3620070}
Jiacheng Ye, Zhiyong Wu, Jiangtao Feng, Tao Yu, and Lingpeng Kong. 2023.
\newblock Compositional exemplars for in-context learning.
\newblock In \emph{Proceedings of the 40th International Conference on Machine
  Learning}, ICML'23. JMLR.org.

\bibitem[{Zellers et~al.(2019)Zellers, Bisk, Farhadi, and
  Choi}]{zellers_recognition_2019}
Rowan Zellers, Yonatan Bisk, Ali Farhadi, and Yejin Choi. 2019.
\newblock From recognition to cognition: Visual commonsense reasoning.
\newblock In \emph{Proceedings of the IEEE/CVF Conference on Computer Vision
  and Pattern Recognition}.

\bibitem[{Zhang et~al.(2020)Zhang, Kishore*, Wu*, Weinberger, and
  Artzi}]{Zhang*2020BERTScore:}
Tianyi Zhang, Varsha Kishore*, Felix Wu*, Kilian~Q. Weinberger, and Yoav Artzi.
  2020.
\newblock \href {https://openreview.net/forum?id=SkeHuCVFDr} {Bertscore:
  Evaluating text generation with bert}.
\newblock In \emph{International Conference on Learning Representations}.

\bibitem[{Zhang et~al.(2019)Zhang, Baldridge, and He}]{zhang-etal-2019-paws}
Yuan Zhang, Jason Baldridge, and Luheng He. 2019.
\newblock \href {https://doi.org/10.18653/v1/N19-1131} {{PAWS}: Paraphrase
  adversaries from word scrambling}.
\newblock In \emph{Proceedings of the 2019 Conference of the North {A}merican
  Chapter of the Association for Computational Linguistics: Human Language
  Technologies, Volume 1 (Long and Short Papers)}, pages 1298--1308,
  Minneapolis, Minnesota. Association for Computational Linguistics.

\bibitem[{Zhao et~al.(2021)Zhao, Wallace, Feng, Klein, and
  Singh}]{pmlr-v139-zhao21c}
Zihao Zhao, Eric Wallace, Shi Feng, Dan Klein, and Sameer Singh. 2021.
\newblock \href {https://proceedings.mlr.press/v139/zhao21c.html} {Calibrate
  before use: Improving few-shot performance of language models}.
\newblock In \emph{Proceedings of the 38th International Conference on Machine
  Learning}, volume 139 of \emph{Proceedings of Machine Learning Research},
  pages 12697--12706. PMLR.

\end{thebibliography}

\clearpage
\appendix

\section{Details of Datasets}
\label{app:data}
The details of all studied datasets are delineated as follows
\begin{itemize}[leftmargin=*]
    \item  \textbf{SNLI Dataset}: The SNLI dataset, a benchmark in natural language inference, encompasses approximately 570,000 human-annotated sentence pairs, each pair formed by a premise and a hypothesis. These sentences originate from an existing corpus of image captions, thus offering a broad spectrum of common subjects and linguistic structures ~\cite{bowman-etal-2015-large}.
    \item \textbf{HANS Dataset}: \citet{mccoy-etal-2019-right} developed a dataset with the express purpose of scrutinizing the performance of models when confronted with sentences characterized by several types of distracting signals. These signals encompass the presence of lexical overlap, sub-sequences, and constituent heuristics between the corresponding hypotheses and premises.
    \item \textbf{Datasets Sensitive to Compositionality (ISCS)}: As proposed by \citet{nie2019analyzing}, a softmax regression model was employed to utilize lexical features present in the premise and hypothesis sentences, thereby generating instances of misclassification. Here, the \emph{Lexically Misleading Score} (LMS) denotes the predicted probability of the misclassified label. Adapting the approach of \citet{liu-etal-2020-empirical}, we concentrated on the subsets possessing LMS values exceeding 0.7.
    \item \textbf{Not another Negation (NaN) NLI Dataset}: NaN dataset is developed to probe the capabilities of NLP models in comprehending sub-clausal negation~\cite{truong-etal-2022-another}.
    \item \textbf{Stress Test Datasets (ST)}: Our analysis also incorporates various stress tests described by \citet{naik-etal-2018-stress} such as ``word overlap'' (ST-WO), ``negation'' (ST-NE), ``length mismatch'' (ST-LM), and ``spelling errors'' (ST-SE). Specifically, ST-WO aims to identify lexical overlap heuristics between the premise and hypothesis, ST-NE seeks to detect intense negative lexical cues in partial-input sentences, ST-LM aspires to create misleading predictions by artificially lengthening the premise using nonsensical phrases, and ST-SE employs spelling errors as a means to deceive the model.
    \item \textbf{Datasets Detected by Classifier (PICD)}: In the approach proposed by \citet{gururangan-etal-2018-annotation}, fastText was applied to hypothesis-only inputs. Subsequent instances from the SNLI test sets \cite{bowman-etal-2015-large} that could not be accurately classified were designated as `hard' instances.
    \item  \textbf{Surface Pattern Datasets (PISP)}: \citet{liu-etal-2020-hyponli} identified surface patterns that exhibit strong correlation with specific labels, thereby proposing adversarial test sets counteracting the implications of surface patterns. As suggested by \citet{liu-etal-2020-empirical}, we employed their `hard' instances extracted from the MultiNLI mismatched development set \cite{williams-etal-2018-broad} as adversarial datasets.

    \item \textbf{Adversarial NLI (ANLI)}: ANLI dataset~\cite{nie-etal-2020-adversarial} is a challenging resource created for training and testing models on NLI, featuring adversarial examples intentionally curated to obfuscate or mislead benchmark models, thereby increasing its challenge factor. This dataset is constructed in multiple rounds, with each subsequent round featuring human-created examples specifically designed to outsmart models trained on the previous rounds. In total, the dataset comprises three distinct rounds, specifically ANLI R1, ANLI R2, and ANLI R3, highlighting the layered complexity of this resource.

    \item \textbf{Quora Question Pairs (QQP)}: QQP dataset~\cite{wang-etal-2018-glue} comprises pairs of questions sourced from the Quora community question-answering platform. The primary objective is to ascertain whether each question pair exhibits semantic equivalence.

    \item  \textbf{Paraphrase Adversaries from Word Scrambling (PAWS)}: The PAWS-QQP dataset~\cite{zhang-etal-2019-paws}, derived from the QQP datasets, targets the intricate task of paraphrasing identification, emphasizing the differentiation of sentences that, despite high lexical similarity, convey distinct meanings. It incorporates adversarial examples generated via word scrambling, presenting a stringent assessment for NLP models. 

\end{itemize}

\section{Meta-prompts for Generating Synthetic NLEs}
\label{app:meta_prompt}
\tabref{tab:meta_prompt} and \ref{tab:meta_qqp_prompt} present the meta-prompts and demonstration instances employed for producing NLEs utilizing ChatGPT in zero- and few-shot scenarios.

\begin{table}[th!]
    \centering
    \begin{tabular}{p{0.9\linewidth}}
    \toprule
      \multicolumn{1}{c}{\bf Meta-prompt for zero-shot generation} \\
      \midrule
         Assume that you're an expert working on natural language inference tasks. Given a premise, a hypothesis, and the corresponding label. Please write a concise and precise reason to explain why the label is assigned to the example:\\
\midrule
\midrule
      \multicolumn{1}{p{0.9\linewidth}}{\bf Meta-prompt and demonstration instances for few-shot generation} \\
      \midrule
Assume that you're an expert working on natural language inference tasks. Given a premise, a hypothesis, and the corresponding label. Please write a concise and precise reason to explain why the label is assigned to the example by following the provided examples: 
\\\\
\textbf{Premise}: A boy peers out of an open window. \\
    \textbf{Hypothesis}: The boy looks out the window. \\
    \textbf{Label}: entailment \\
    \textbf{NLE}: The boy peers out of a window, so the boy looks out the window.\\
===== \\
    \textbf{Premise}: A kid doing a trick on a skateboard. \\
    \textbf{Hypothesis}: The kid eating lunch inside the cafeteria. \\
    \textbf{Label}: contradiction \\
   \textbf{NLE}: The kid cannot be doing a trick and eating lunch at the same time \\
===== \\
    \textbf{Premise}: A man jumps off of his skateboard on the top of a cement ramp. \\
    \textbf{Hypothesis}: a man jumps off a skateboard at the top of a ramp.\\
    \textbf{Label}: neutral\\
    \textbf{NLE}: A man can jump off a skateboard without being at the top of a ramp.\\
         \bottomrule
    \end{tabular}
    \caption{Meta-prompts used to generate NLEs via ChatGPT in zero- and few-shot scenarios for natural language inference tasks.}
    \label{tab:meta_prompt}
\end{table}

\begin{table}[th!]
    \centering
    \begin{tabular}{p{0.9\linewidth}}
    \toprule
      \multicolumn{1}{c}{\bf Meta-prompt for zero-shot generation} \\
      \midrule
         Assume that you're an expert working on paraphrasing identification tasks. Given two sentences and the corresponding label. Please write a concise and precise reason to explain why the label is assigned to the example:\\
\midrule
\midrule
     \multicolumn{1}{p{0.9\linewidth}}{\bf Meta-prompt and demonstration instances for few-shot generation} \\
      \midrule
Assume that you're an expert working on paraphrasing identification tasks. Given two sentences and the corresponding label. Please write a concise and precise reason to explain why the label is assigned to the example by following the provided examples: \\
\\
\textbf{Q1}: Does life get harder as you get older?\\
\textbf{Q2}: Does life really get harder as you get older?\\
\textbf{Label}: duplicate\\
\textbf{NLE}: Both questions ask whether life does get harder as you get older. \\
===== \\
\textbf{Q1}: What is the National nanotechnology initiative?\\
\textbf{Q2}: What is the lead time for SSN4EGS411 board?\\
\textbf{Label}: not duplicate\\
\textbf{NLE}: completely different questions
\\
         \bottomrule
    \end{tabular}
    \caption{Meta-prompts used to generate NLEs via ChatGPT in zero- and few-shot scenarios for paraphrasing identification tasks.}
    \label{tab:meta_qqp_prompt}
\end{table}

\section{Supplementary Studies}
\label{app:suppl}

\paragraph{Using NLEs Generated by Vicuna and Llama2.} Our research demonstrates that the integration of NLEs generated by ChatGPT significantly enhances the performance of \xicl for five advanced LLMs. To assess the efficacy of these ChatGPT-generated NLEs, we explore the generation of synthetic NLEs using Vicuna and Llama2, ranked as the third and second-best models respectively. Likewise, these NLEs are generated in a few-shot setting, referred to herein as Vicuna$_{\text{few}}$ and Llama2$_{\text{few}}$, respectively. To ensure a fair comparison, we employ Vicuna as the underlying model to evaluate \casesix, \casefive, and \casethree on all studied datasets.

\begin{table}[h]
    \centering
    \scalebox{0.8}{
    \begin{tabular}{cccc}
    \toprule
     \multirow{2}{*}{\textbf{Tasks}} &  \multicolumn{3}{c}{\textbf{NLEs}}\\
     \cmidrule(lr){2-4}
     &  \textbf{ fs-Vicuna}& \textbf{ fs-Llama2}&\textbf{fs-ChatGPT}  \\
    \midrule
SNLI & 62.9 (\ \  \Minus 5.0) & 64.1 (\ \  \Minus 3.7)& 65.0 (\ \  \Minus 2.9)\\
HANS &55.5 (\ \ \Minus 7.4) & 67.4 (\ \ \Plus4.5) & 74.5 (\Plus 11.6)\\
ISCS  &65.1 (\ \ \Plus 4.2) & 63.6 (\ \ \Plus2.7) & 65.5 (\ \ \Plus 4.6)\\
NaN  & 62.6 (\ \ \Minus 1.6) & 65.1 (\ \ \Plus0.9) & 66.3 (\ \ \Plus 2.1)\\
ST  &59.5 (\ \ \Plus 2.2) & 61.9 (\ \ \Plus4.6)
 & 64.8 (\ \ \Plus 7.5)\\
PICD & 60.2 (\ \ \Minus 3.5) & 60.8 (\ \ \Minus 2.9)& 61.6 (\ \ \Minus 2.1)\\
PISP & 66.0 (\Plus 11.0) & 66.1 (\Plus11.1)& 66.0 (\Plus 11.0) \\
ANLI (R1) & 66.1 (\ \ \Plus 9.1) & 65.8 (\ \ \Plus8.8)& 64.9 (\ \ \Plus 7.9)\\
ANLI (R2) &  55.4 (\ \ \Plus 6.5) &55.9 (\ \ \Plus7.0) & 55.5 (\ \ \Plus 6.6)\\
ANLI (R3) & 49.6 (\Plus 10.8) & 50.7 (\Plus11.9)& 52.0 (\Plus 13.2)\\		
\midrule
Average & 60.3 (\ \ \Plus 3.8) & 62.1 (\ \ \Plus 5.6)	& \textbf{63.5} (\ \ \Plus 6.9) \\
    \bottomrule
    \end{tabular}}
    \caption{ICL performance of Vicuna using (1) standard ICL without NLEs, (2) \xicl with Vicuna-generated NLEs in a few-shot scenario: fs-Vicuna, (3) \xicl with Llama2-generated NLEs in a few-shot scenario: fs-Llama2, (4) \xicl with ChatGPT-generated NLEs in a few-shot scenario: fs-ChatGPT. Numbers in the parentheses represent differences compared to \casetwo.}
    \label{tab:llama}
    \vspace{-0.5cm}
\end{table}

Our results, detailed in \cref{tab:llama}, highlight that \xicl generally gains greater benefit from LLM-generated NLEs as opposed to those produced by humans. Meanwhile, \casethree consistently outperforms \casesix and \casefive considerably, except for ANLI R1 and R2. These findings suggest that to harness the potential of AI-generated NLEs fully, the employment of a powerful LLM is integral.

\begin{figure*}[]
    \centering
    \includegraphics[width=0.99\textwidth]{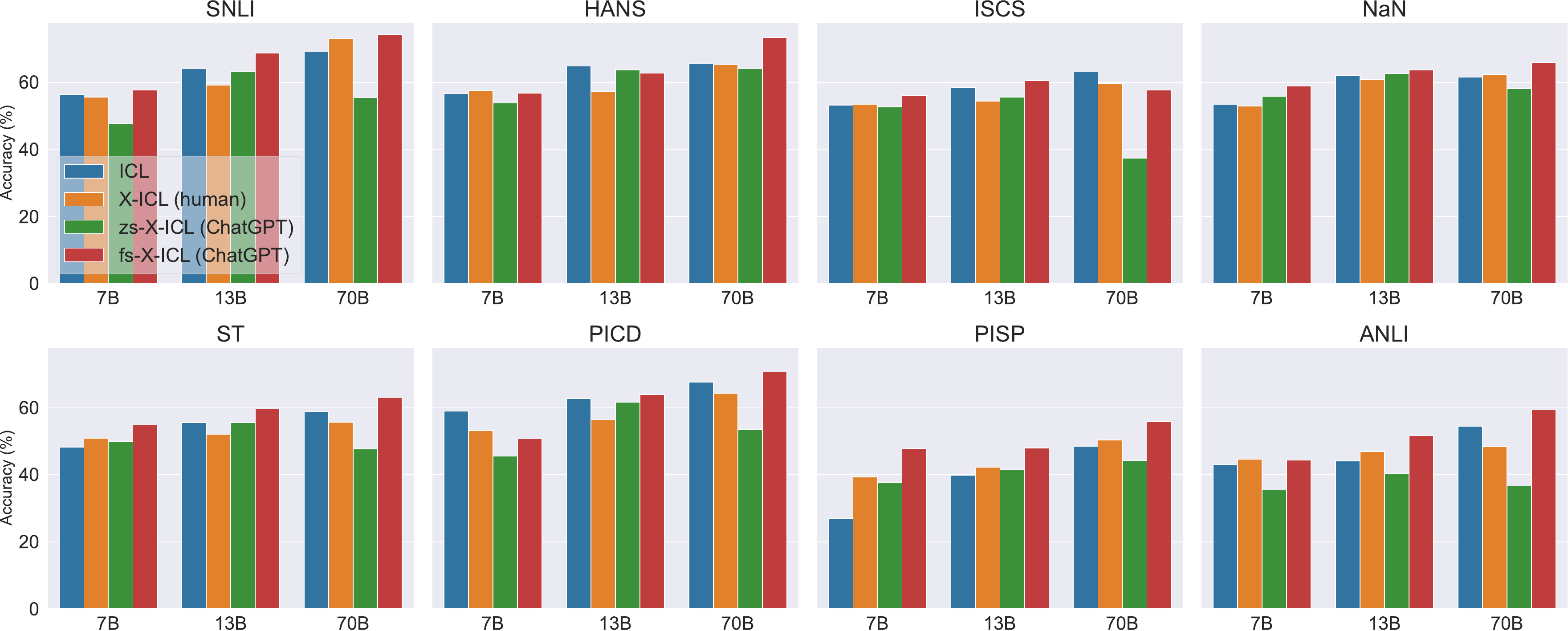}
    \caption{ICL performance of Llama2 (7B, 13B, 70B) using (1) standard ICL without NLEs, (2) \xicl with human-written NLEs: \casetwo, (3) \xicl with ChatGPT-generated NLEs in a zero-shot scenario: \casefour, (4) \xicl with ChatGPT-generated NLEs in a few-shot scenario:\casethree. ANLI is the average of R1, R2 and R3.}
    \label{fig:diff_model}
\end{figure*}

\begin{table*}[t]
    \centering
   \scalebox {0.8} {
    \begin{tabular}{lccc|ccc|ccc|ccc}
    \toprule
      & \multicolumn{3}{c}{\textbf{NaN}} & \multicolumn{3}{c}{\textbf{PICD}} & \multicolumn{3}{c}{\textbf{ANLI (R1)}} & \multicolumn{3}{c}{\textbf{ANLI (R2)}}\\
      \cmidrule{2-13}
      & e-SNLI & ANLI & $\lvert \Delta \rvert$ &  e-SNLI & ANLI & $\lvert \Delta \rvert$ &  e-SNLI & ANLI & $\lvert \Delta \rvert$ &  e-SNLI & ANLI & $\lvert \Delta \rvert$ \\
      \midrule
    \caseone & 70.0& 69.4 & 0.6 & 64.0& 64.1 & 0.1 &  52.6 &  62.4 & 9.7 & 43.9  & 51.7& 7.8 \\
\casethree& 73.1 &71.8 & 1.2 & 76.9 & 76.1 & 0.8 & 65.0 & 68.5 & 3.5 & 53.2 & 54.4 & 1.2\\
    \bottomrule
    \end{tabular}
    }
    \caption{Performance of \caseone and \casethree employing e-SNLI and ANLI as prompts for testing NaN, PICD, and ANLI (R1/R2). $\lvert \Delta \rvert$ signifies the absolute difference in the performance outcomes when utilizing e-SNLI in contrast to ANLI. The backbone model is GPT3.5-turbo.}
    \label{tab:cross}
\end{table*}

\paragraph{Does model size matter?} 
{We have shown the efficacy of \xicl across a range of LLMs of varying sizes. However, the variability in data and training processes among these models renders the applicability of our approach to smaller-scale models inconclusive, especially since the smaller models often exhibit less benefit from NLEs compared to larger models within the same family~\cite{wei2022emergent}. Therefore, we have evaluated our approach using three distinct sizes of Llama2 models: 7B, 13B, and 70B parameters.}

Referring to \figref{fig:diff_model}, one can find the performance of both ICL and \xicl generally improves in correspondence with the escalation of model size, except for \casefour. Moreover, the gap in performance between \caseone and \casethree widens, indicating that models with greater capabilities derive increased benefits from NLEs. This observation aligns with the results reported by~\citet{wei2022emergent}.

\paragraph{Distribution Shift Prompting.} Previous works indicate that \xicl can potentially encourage LLMs to engage in deliberate thinking, a predominant factor responsible for substantial performance improvements over the standard \caseone in complex reasoning tasks~\cite{wei2022chain}. In addition, our findings have demonstrated a dramatic enhancement in the robustness of LLMs due to \xicl, which contributes to significant improvements in \caseone when applied to various adversarial datasets.

Moreover, a previous study established that upon understanding the concept underlying particular tasks, humans can address similar tasks despite a distribution shift~\cite{d69ed548-5302-3d45-87a7-4fea0bc35a5e}. To explore the robustness of \caseone and \xicl against distribution shifts, we employ the e-SNLI dataset as the demonstration set for ANLI (R1/R2), while utilizing the ANLI training set for testing NaN and PICD. Due to its outstanding performance, we use GPT3.5-turbo as the backbone model.

As suggested in \tabref{tab:cross}, for NaN and PICD, using e-SNLI as the prompt proves to be more effective than ANLI for both \caseone and \casethree. This improvement can be attributed to the distribution shift.
Likewise, the distribution shift results in a noticeable distinction between e-SNLI and ANLI for \caseone on ANLI (R1/R2). Nonetheless, incorporating NLEs enables \casethree to substantially reduce this gap, from 9.7 to 3.5 for ANLI (R1), and from 7.8 to 1.2 for ANLI (R2). This finding indicates that \xicl may improve the robustness of LLMs in the face of distribution shifts.

\paragraph{Analysis on memorization} LLMs such as ChatGPT have occasionally replicated instances from renowned benchmark datasets, including MNLI and BoolQ~\cite{sainz2023}. This unintentional \textit{`contamination'} might contribute to misconceptions regarding the superior performance of LLMs on these widespread benchmarks due to data memorization.

Following~\citet{carlini2023quantifying}, we merge the premise and hypothesis of each test instance into a single sentence, using the first part as the prefix. If an LLM could perfectly replicate the second part, we labeled the instance as \textit{`extractable'}. Evaluating all studied models, we observe that the proportion of extractable instances is under 0.001\%
across all datasets and backbone models, indicating that the superior performance of LLMs might not be ascribed to memorization.

\section{Qualitative Analysis on NLEs}
\subsection{Qualitative Analysis on NLEs for Demonstration Set}
\label{app:qual_analysis_demo}
We first conducted a qualitative analysis of NLEs generated by ChatGPT under zero- and few-shot scenarios, using the demonstration set as a basis. Note that each instance in the demonstration set has three distinct NLEs: (1) the zero-shot NLE from ChatGPT, (2) the few-shot NLE from ChatGPT, and (3) the human-written NLE. From these three NLEs per instance, one was randomly selected, and both the instance and the chosen NLE were incorporated into the evaluation set.

Subsequently, this evaluation set was rated independently by four authors on a 5-point Likert scale to assess the quality of the NLEs. The scale ranges were 1 (extremely dissatisfied), 2 (dissatisfied), 3 (neutral), 4 (satisfied), and 5 (extremely satisfied). Finally, we calculated the average scores for both ChatGPT-generated and human-written NLEs for each evaluator.

\subsection{Qualitative Analysis on NLEs for Inference Set}
\label{app:qual_analysis_test}
We also conducted a qualitative analysis of NLEs generated by \casethree, utilizing GPT3.5-turbo as the foundational model. A total of 280 randomly sampled, correctly predicted examples from \casethree were distributed evenly among seven evaluators. These evaluators were tasked to assess the quality of the NLE for each assigned instance, based on the premise-hypothesis pair and its corresponding correctly predicted label.

The evaluators were required to rate the quality of the NLE using the aforementioned 5-point Likert scale. In case of dissatisfaction, they were asked to identify the reason from a list of predefined factors, including:
\begin{itemize}
    \item \textbf{template}: The NLE simply restates the input and employs it as a justification.
    \item \textbf{insufficient justification}: The NLE requires more support for the prediction.
    \item \textbf{too verbose}: The NLE is overly detailed and includes unnecessary information.
    \item \textbf{incorrect arguments}: Despite the prediction being accurate, the NLE fails to support it due to erroneous arguments.
    \item \textbf{contradict commonsense}: The NLE is incorrect and contradicts commonsense.
    \item \textbf{hallucinations}: The NLE includes fabricated information.
\end{itemize}

\begin{figure}[t]
    \begin{subfigure}[b]{0.95\linewidth}
         \centering
         \includegraphics[width=0.95\linewidth]{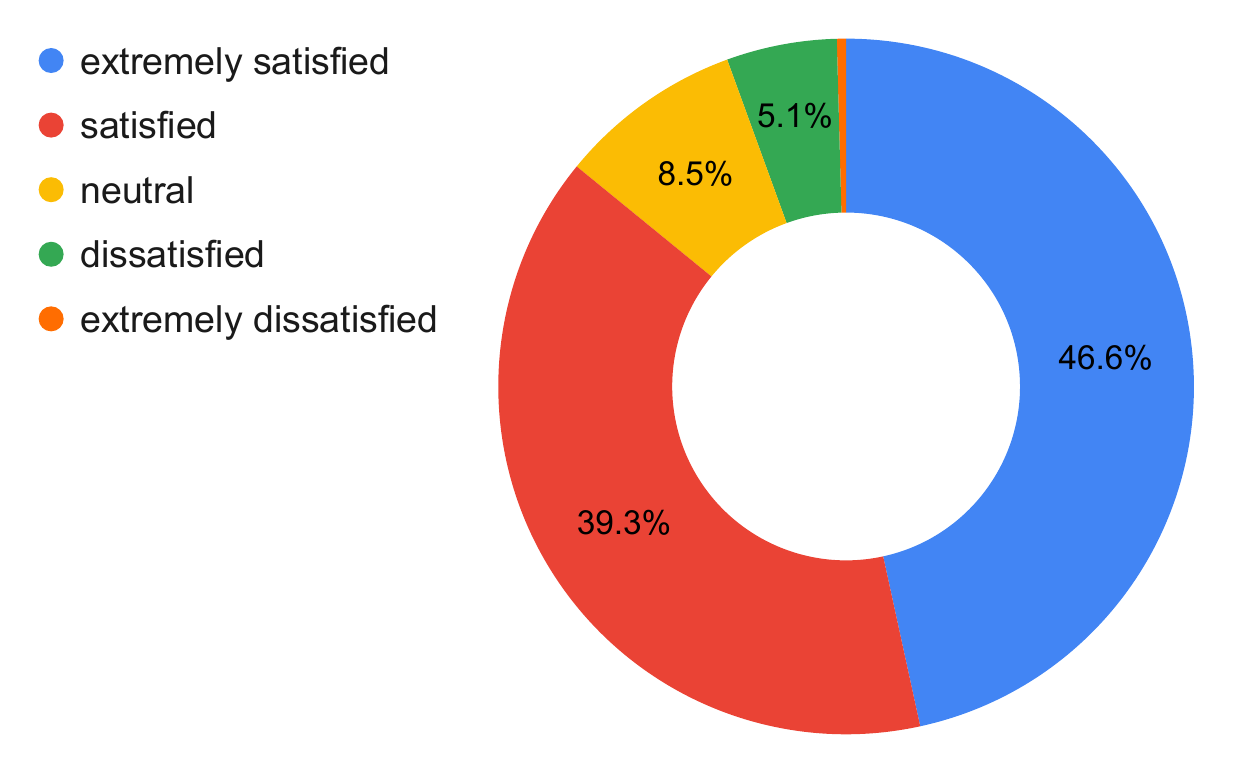}
     \end{subfigure}
     \par\bigskip
   \begin{subfigure}[b]{0.95\linewidth}
         \centering
         \includegraphics[width=0.95\linewidth]{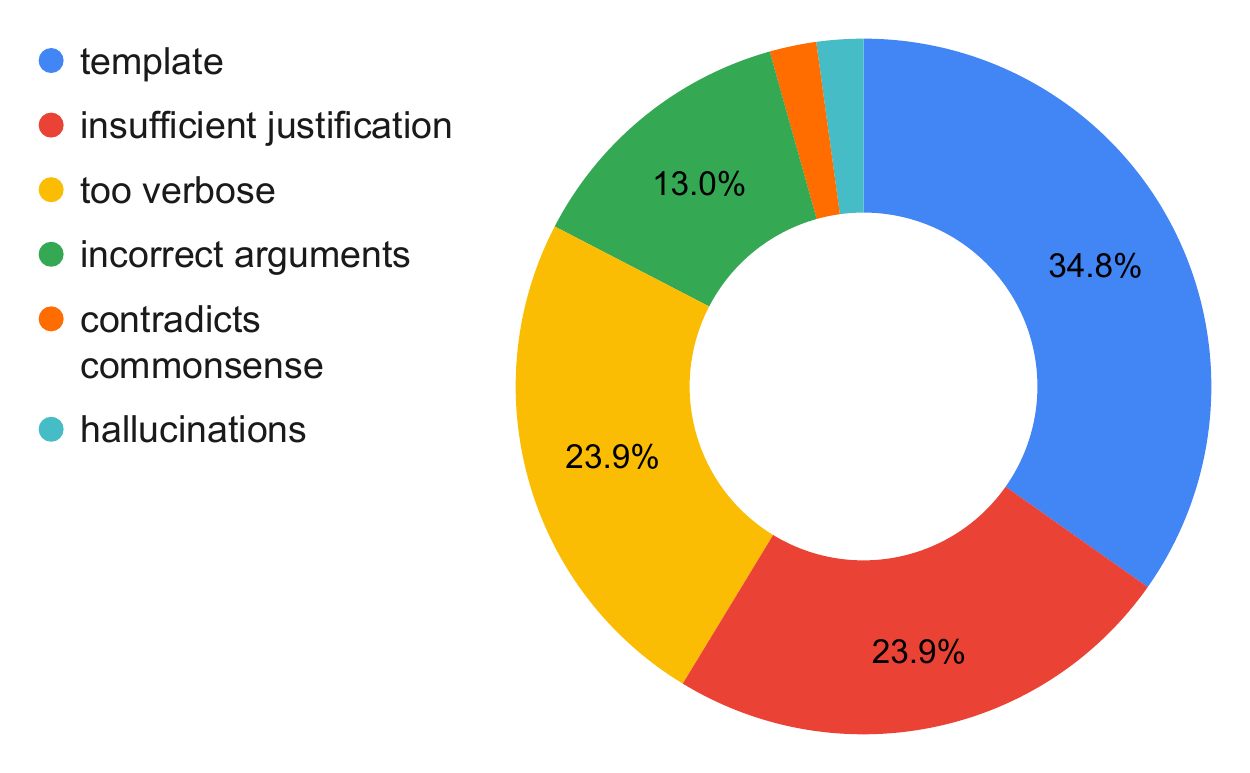}
     \end{subfigure}
     \caption{Human evaluation on ChatGPT-generated NLEs for the correct predictions from \casethree. \textbf{Top}: distribution of satisfaction scores. \textbf{Bottom}: distribution of reasons for dissatisfaction.}
     \label{fig:nle_quality}
\end{figure}

According to~\cref{fig:nle_quality}, 46.6\% and 39.3\% of NLEs are marked as `extremely satisfied' and `satisfied' respectively, constituting 85.9\% of the total 280 NLE samples. This suggests a high-quality output from GPT3.5-turbo in general. As for the lower-quality NLEs, the primary reasons for dissatisfaction include `template', `insufficient justification', and `too verbose'. Interestingly, this suggests that, despite the expressed dissatisfaction, evaluators generally did not find incorrect justifications in most instances.

\begin{table*}[th!]
    \centering
    \begin{tabular}{p{0.9\linewidth}}
    \toprule
      \textbf{Premise}: The organizing committee said 65 countries have entered the Lillehammer Olympic Games, matching the number of nations at the 1992 Winter Games in Albertville, France.\\
\textbf{Hypothesis}: The Lillehammer Olympic games take place 12 years after the Albertville Olympic games.\\
\textbf{NLE}: The premise does not provide any information about the time difference between the Lillehammer Olympic Games and the Albertville Olympic Games. Therefore, there is no evidence to support or contradict the hypothesis, resulting in a neutral label.\\
\textbf{Predicted Label}: neutral\\
\textbf{Correct Label}: neutral\\
\midrule
\textbf{Premise}: A woman was not injured when she was carjacked in west Charlotte Tuesday morning. According to Charlotte-Mecklenburg police, two men reportedly pulled a gun on a woman who was stopped at a stop sign on State Street at Whitehaven Avenue around 3:30 a.m. Police say the men then stole the woman's Honda Fit. Officers have not found the stolen vehicle. No other details were released. Copyright 2018 WBTV. All rights reserved.\\
\textbf{Hypothesis}: A woman stole a Honda Fit in west Charlotte at 3:30 AM but was not injured.\\
\textbf{NLE}: The premise states that a woman was carjacked in west Charlotte at 3:30 AM and that two men stole her Honda Fit. The hypothesis suggests that the woman stole a Honda Fit, which contradicts the information provided in the premise. Therefore, the hypothesis is false and leads to a contradiction.\\
\textbf{Predicted Label}: contradiction\\
\textbf{Correct Label}: contradiction\\
\midrule
\textbf{Premise}: New question. Last question. For you, Governor. And this flows somewhat out of the Boston debate. You, your running mate, your campaign officials have charged that Vice President Gore exaggerates, embellishes and stretches the facts, etcetera. Are you -- do you believe these are serious issues? This is a serious issue that the voters should use in deciding which one of you two men to vote for on November 7?\\
\textbf{Hypothesis}: the Vice President exaggerates the truth.\\
\textbf{NLE}: The premise states that the question is about whether the Governor believes that the Vice President exaggerates, embellishes, and stretches the facts. Therefore, the hypothesis is a restatement of the question and can be confirmed as true based on the information provided in the premise.\\
\textbf{Predicted Label}: entailment\\
\textbf{Correct Label}: entailment\\
\midrule
\textbf{Premise}: Budapest consists of two parts, Buda and Pest, which are situated on opposite sides of the river and connected by a series of bridges.\\
\textbf{Hypothesis}: Buda and Pest function as separate cities.\\
\textbf{NLE}: The premise states that Budapest consists of two parts, Buda and Pest, which are connected by a series of bridges. However, it does not suggest that Buda and Pest function as separate cities. Therefore, the hypothesis cannot be confirmed or refuted based on the given information, resulting in a neutral label.\\
\textbf{Predicted Label}: neutral\\
\textbf{Correct Label}: neutral\\
         \bottomrule
    \end{tabular}
    \caption{Test instances and the corresponding generated NLEs from GPT-3.5-turbo for the correct predictions.}
    \label{tab:chatgpt-quality}
\end{table*}

\begin{table*}[th!]
    \centering
    \begin{tabular}{p{0.9\linewidth}}
    \toprule
      \textbf{Premise}: Mulberry () is a 1986 South Korean drama film directed by Lee Doo-yong. Based on the eponymous classic story by Na Do-hyang, the film became known for its erotic subject matter, made possible by the government's gradual relaxation of censorship and control over the film industry during the 1980s. It was part of the wave of "Folk erotic" films that had become popular in South Korea at this time.\\
\textbf{Hypothesis}: Mulberry () is a 1986 South Korean drama film directed by Lee Doo-dong.\\
\textbf{NLE}: The hypothesis is identical to the premise, except for a minor spelling error in the director's name. Therefore, the hypothesis is true based on the provided premise.\\
\textbf{Predicted Label}: entailment\\
\textbf{Correct Label}: contradiction\\
\midrule
\textbf{Premise}: In the Company of Men is a 1997 Canadian/American black comedy written and directed by Neil LaBute and starring Aaron Eckhart, Matt Malloy, and Stacy Edwards. The film, which was adapted from a play written by LaBute, and served as his feature film debut, won him the Independent Spirit Award for Best First Screenplay.\\
\textbf{Hypothesis}: Neil LaBute worked with Aaron Eckhart for the comedy that was his first screenplay.\\
\textbf{NLE}: The premise states that Neil LaBute wrote and directed In the Company of Men, which was his feature film debut and won him the Independent Spirit Award for Best First Screenplay. It also mentions that the film starred Aaron Eckhart, but it does not provide any information about whether or not LaBute worked with Eckhart specifically for this film. Therefore, the hypothesis cannot be either true or false based on the provided premise.\\
\textbf{Predicted Label}: neutral\\
\textbf{Correct Label}: entailment\\
\midrule
\textbf{Premise}: Barry Cunningham (born 19 December 1952) is a British publisher, who worked for various publishers including Penguin Books and Bloomsbury before setting up Chicken House publishing in 2000. He is most well known for signing J.K Rowling, and publishing "Harry Potter and the Philosopher’s Stone" in 1997.\\
\textbf{Hypothesis}: Barry Cunningham is most proud of signing J.K Rowling, and publishing "Harry Potter and the Philosopher’s Stone" in 1997.\\
\textbf{NLE}: The premise states that Neil LaBute wrote and directed In the Company of Men, which was his feature film debut and won him the Independent Spirit Award for Best First Screenplay. It also mentions that the film starred Aaron Eckhart, but it does not provide any information about whether or not LaBute worked with Eckhart specifically for this film. Therefore, the hypothesis cannot be either true or false based on the provided premise.\\
\textbf{Predicted Label}: entailment\\
\textbf{Correct Label}: neutral\\
\midrule
\textbf{Premise}: Gyula Trebitsch (3 November 1914 - 12 December 2005) was a German film producer born in Budapest, Hungary. He was nominated in 1956 for the Academy Award for Best Foreign Language Film along with Walter Koppel for their film "The Captain of Kopenick".\\
\textbf{Hypothesis}: Gyula Trebitsch was nominated for the Academy Award for Best Foreign Language Film for his work on "The Captain of Kopenick" at the age of 43.\\
\textbf{NLE}: The premise states that Gyula Trebitsch was nominated for the Academy Award for Best Foreign Language Film for "The Captain of Kopenick" in 1956, but it does not provide information about his age at the time of the nomination. Therefore, the hypothesis cannot be
 either true or false based on the provided premise.\\
\textbf{Predicted Label}: neutral\\
\textbf{Correct Label}: contradiction\\
         \bottomrule
    \end{tabular}
    \caption{Test instances and the corresponding generated NLEs from GPT-3.5-turbo for the incorrect predictions.}
    \label{tab:chatgpt-quality-incorrect}
\end{table*}



\section{Human-written NLEs for QQP}
\label{app:qqp}
Given the absence of NLEs for the 
QQP dataset, we have conducted a randomized sampling of 32 instances from the QQP training set. For each selected instance, we crafted a corresponding NLE. The details of these QQP instances and their respective NLEs are presented in \tabref{tab:qqp_nle1}, \ref{tab:qqp_nle2}, \ref{tab:qqp_nle3}, and \ref{tab:qqp_nle4}.

\begin{table*}[th!]
    \centering
    \begin{tabular}{p{0.9\linewidth}}
    \toprule
      \textbf{Q1}: Is 6 foot 4 too tall as an ideal height for a man?\\
\textbf{Q2}: My height is 5'6 and I'm 14 year old boy, my mom is 5'4 and my dad is 5'7. How tall will I be?\\
\textbf{Label}: not duplicate\\
\textbf{NLE}: Predicting future height given parents' heights concerns genetic factors of height, whereas ideal height for man concerns more about its social aspect.\\
\midrule
\textbf{Q1}: Approximately how many hours have you spent on the internet till date?\\
\textbf{Q2}: What amount of time do you spent on the Internet?\\
\textbf{Label}: not duplicate\\
\textbf{NLE}: Total number of hours spend on Internet till date not just depend on the average hours on internet per day, but also many other factors such as the age the user started using it.\\
\midrule
\textbf{Q1}: What are the most ridiculous statements made by Donald Trump?	\\
\textbf{Q2}: My black friend supports Donald Trump, is that ridiculous?\\
\textbf{Label}: not duplicate\\
\textbf{NLE}: Asking the most ridiculous statement made by Donald Trump is different than asking why a supporter support him. A supporter can support him for other reasons.\\
\midrule
\textbf{Q1}: "What is the origin of the phrase ""pipe dream""?"\\
\textbf{Q2}: "How did the phrase ""toe head"" originate?"\\
\textbf{Label}: not duplicate\\
\textbf{NLE}: The two questions asked about the origin of two different words.\\
\midrule
\textbf{Q1}: What is a good first programming language to learn?\\
\textbf{Q2}: What is the most valuable programming language for the future to learn?\\
\textbf{Label}: duplicate\\
\textbf{NLE}: When picking a good first programming language to learn, people may consider the most valuable one language if they learn it for making money.\\
\midrule
\textbf{Q1}: What is best way for earning money?	\\
\textbf{Q2}: How can I start making money? What are the best ways to make money?	\\
\textbf{Label}: duplicate\\
\textbf{NLE}: Both questions ask about what are best ways to make money\\
\midrule
\textbf{Q1}: Does the Indian education system need a reformation?\\
\textbf{Q2}: Should the education system be changed in India? If so why or why not?\\
\textbf{Label}: duplicate\\
\textbf{NLE}: Both questions essentially inquire about the necessity and justification for changing the Indian education system.\\
\midrule
\textbf{Q1}: What is the application of quantum physics?\\
\textbf{Q2}: What are some applications of quantum physics?\\
\textbf{Label}: duplicate\\
\textbf{NLE}: The two questions both seek information about the practical use of quantum physics.\\
         \bottomrule
    \end{tabular}
    \caption{QQP instances and the corresponding NLEs.}
    \label{tab:qqp_nle1}
\end{table*}

\begin{table*}[th!]
    \centering
    \begin{tabular}{p{0.9\linewidth}}
    \toprule
    \textbf{Q1}: How is the word 'calumny' used in a sentence?\\
\textbf{Q2}: How is the word 'mischievous' used in a sentence?\\
\textbf{Label}: not duplicate\\
\textbf{NLE}: The two questions ask about two different words with different meanings.\\
\midrule
\textbf{Q1}: What are your views on the abolishment of 500 rupees note?\\
\textbf{Q2}: How will the ban of Rs 500 and Rs 1000 notes affect Indian economy?\\	
\textbf{Label}: not duplicate\\
\textbf{NLE}: The former question asks specifically about the abolishment of the Rs 500 note, while the latter asks about the Rs 500 and the Rs 1000 notes.\\
\midrule
\textbf{Q1}: What are the valence electrons of titanium?\\
\textbf{Q2}: What is the number of valence electrons in hydrogen? How is this determined?\\
\textbf{Label}: not duplicate\\
\textbf{NLE}: The former question asks about titanium, while the latter is about hydrogen.\\
\midrule
\textbf{Q1}: Do movie actors get paid each time their movie is played on TV?\\
\textbf{Q2}: Why are film actors so highly paid whereas scientists are paid relatively quite little?\\
\textbf{Label}: not duplicate\\
\textbf{NLE}: The former question asks some details about how actors get paid, while the latter asks about the gap between actor and scientist salaries.\\
\midrule
\textbf{Q1}: How do I build an electromagnetic propulsion engine?\\
\textbf{Q2}: How would I build a magnetic propulsion system?\\
\textbf{Label}: duplicate\\
\textbf{NLE}: Both question asks about building magnetic propulsion systems.\\
\midrule
\textbf{Q1}: Why is salt water taffy candy imported in France?\\
\textbf{Q2}: Why is Saltwater taffy candy imported in The Bahamas?	\\
\textbf{Label}: duplicate\\
\textbf{NLE}: Both questions ask about the reasons behind importing salt water taffy candy.\\
\midrule
\textbf{Q1}: Why do we call Java platform independent language when it still requires platform dependent JVM to get executed?\\
\textbf{Q2}: How is the Java platform independent when we need to have JVM on every machine to run Java programs?\\
\textbf{Label}: duplicate\\
\textbf{NLE}: Both questions ask why do we call Java platform-independent, since it still depends on the availability of a JVM.\\
\midrule

\textbf{Q1}: What are the various ways through which one can earn money online?\\
\textbf{Q2}: How do you make easy money online?\\
\textbf{Label}: duplicate\\
\textbf{NLE}: Both questions ask how to make money online.\\
         \bottomrule
    \end{tabular}
    \caption{QQP instances and the corresponding NLEs.}
    \label{tab:qqp_nle2}
\end{table*}

\begin{table*}[th!]
    \centering
    \begin{tabular}{p{0.9\linewidth}}
    \toprule
   \textbf{Q1}: Why can't some people think for themselves?\\
\textbf{Q2}: Why don't people think for themselves?\\
\textbf{Label}: not duplicate\\
\textbf{NLE}: "some people" means not all people as the second question seems to imply\\
\midrule
\textbf{Q1}: Why don't we use Solar Furnace to produce electricity?\\
\textbf{Q2}: Why don't we make Solar Cars?\\
\textbf{Label}: not duplicate\\
\textbf{NLE}: using Solar Furnace you can produce some amount of electricity but it may not enough to power a whole car\\
\midrule
\textbf{Q1}: What is an intuitive explanation of the fractional quantum Hall effect?\\
\textbf{Q2}: What is an intuitive explanation of the Quantum Hall effect?\\
\textbf{Label}: not duplicate\\
\textbf{NLE}: fractional quantum Hall effect is different than the Quantum Hall effect, which refers to the integer quantum Hall effect\\
\midrule
\textbf{Q1}: Can INTPs become successful entrepreneurs?\\
\textbf{Q2}: I am business associate in tcs?\\
\textbf{Label}: not duplicate\\
\textbf{NLE}: completely different questions\\
\midrule
\textbf{Q1}: How can I be like Sheldon Cooper?\\
\textbf{Q2}: How do I become like Sheldon Cooper?\\
\textbf{Label}: duplicate\\
\textbf{NLE}: "be like" and "become like" someone is the same thing\\
\midrule
\textbf{Q1}: What do people think about Anonymous?\\
\textbf{Q2}: What do you think about the 'Anonymous' option on Quora?\\
\textbf{Label}: duplicate\\
\textbf{NLE}: "what do people think" and "what do you think" are usually used interchangeably\\
\midrule
\textbf{Q1}: What's the meaning of life?\\
\textbf{Q2}: "What is the meaning of ""Life""?"\\
\textbf{Label}: duplicate\\
\textbf{NLE}: same question with minor different spellings\\
\midrule
\textbf{Q1}: What is it in for the Ibibo group employees with the Makemytrip merger / Buyout?\\
\textbf{Q2}: How do Ibibo employees feel about MakeMyTrip acquiring Ibibo?\\
\textbf{Label}: duplicate\\
\textbf{NLE}: "the Makemytrip merger / Buyout" refers to "MakeMyTrip acquiring Ibibo" and "what is it in for the employees" means "how do the employees feel about"\\
         \bottomrule
    \end{tabular}
    \caption{QQP instances and the corresponding NLEs.}
    \label{tab:qqp_nle3}
\end{table*}

\begin{table*}[th!]
    \centering
    \begin{tabular}{p{0.9\linewidth}}
    \toprule
   \textbf{Q1}: Why is Lionel Messi so brilliant?\\
\textbf{Q2}: Is Lionel Messi a genius?\\
\textbf{Label}: not duplicate\\
\textbf{NLE}: the first question asks for the reason, while the second question inquires about yes or no\\
\midrule
\textbf{Q1}: What are some of the best CyanogenMod 12.1 themes?
\\
\textbf{Q2}: How do I make my own cyanogen 12.1 themes?\\
\textbf{Label}: one asks for the best, whereas the other asks for how\\
\midrule
\textbf{Q1}: Study tips to pas ca ipcc?\\
\textbf{Q2}: If you are unhappy with your current job, would you quit right away \& find another job or wait until you find a job. What are the pros \& cons of each?\\
\textbf{Label}: not duplicate\\
\textbf{NLE}: completely different questions\\
\midrule
\textbf{Q1}: How long does Klonopin (Clonazepam) stay in your system?\\
\textbf{Q2}: How long does 1 mg of Klonopin keep working in your system?\\
\textbf{Label}: not duplicate\\
\textbf{NLE}: the second question gives the exact amount, but the first question doesn't\\
\midrule
\textbf{Q1}: Is a third World War imminent?\\
\textbf{Q2}: How close is a World War III?\\
\textbf{Label}: duplicate\\
\textbf{NLE}: "imminent" means will happen very soon, which is equivalent to "close"\\
\midrule
\textbf{Q1}: What are some of the resources to learn about IoT?\\
\textbf{Q2}: What are the best resources to learn about the Internet of Things (IoT)?\\
\textbf{Label}: duplicate\\
\textbf{NLE}: both ask for the resources for IoT\\
\midrule
\textbf{Q1}:Which are some of the best movies of 2016?\\
\textbf{Q2}: What has been the best movie of 2016?\\
\textbf{Label}: duplicate\\
\textbf{NLE}: both ask for the best movie of 2016\\
\midrule
\textbf{Q1}: Why is Saltwater taffy candy imported in Switzerland?\\
\textbf{Q2}: Why is Saltwater taffy candy imported in the Philippines?\\
\textbf{Label}: duplicate\\
\textbf{NLE}: both ask for the import of Saltwater taffy candy, albeit the different locations\\
         \bottomrule
    \end{tabular}
    \caption{QQP instances and the corresponding NLEs.}
    \label{tab:qqp_nle4}
\end{table*}

   

\end{document}